%% file: iclr2027_conference.tex
\documentclass{article} 
\usepackage{iclr2027_conference,times}

\input{math_commands.tex}

\usepackage{hyperref}
\usepackage{url}
\usepackage{graphicx}
\usepackage{booktabs}
\usepackage{xcolor}
\usepackage{colortbl}
\usepackage{arydshln}
\usepackage{amsmath}
\usepackage{multirow}

\title{KVCMAS: Efficient KV cache Correction \\ for Shared Context in Multi-Agent Systems}

\author{Hyesung Jeon, Hyeongju Ha, Seoyoung Lee, Beomseok Kang, Jae-Joon Kim\\
Department of Electrical and Computer Engineering\\
Seoul National University\\
\texttt{\{hjeon2k,mnv1009,seoyoung.lee29,beomseok,kimjaejoon\}@snu.ac.kr} \\
}

\iclrfinalcopy 
\begin{document}

\maketitle

\begin{abstract}
Prompt-specialized multi-agent systems enable multiple agents to share a model while performing complementary roles to solve complex tasks.
However, agent-specific prefixes change the KV cache generated for the same shared context, causing each agent to repeatedly prefill the growing context and construct a separate cache with high computation and memory overhead.
Selective recomputation reduces this redundancy but still retains substantial model execution, while existing delta correction methods either support only recurring context relations or maintain memory-intensive online correction states for dynamically changing context.
For first seen shared context, these methods also construct a reference cache outside the agent workflow, and an approximate correction at the first agent affects the outputs passed to subsequent agents.
We present \textbf{KVCMAS}, an online KV cache correction framework that represents cross-agent cache deviations using compact low-rank states and seamlessly chains corrections along the agent workflow without an additional reference prefill.
This design supports dynamically changing shared context while preserving an exact first-agent cache.
Across multiple language and vision-language workloads, KVCMAS matches or improves the accuracy of prior KV cache sharing methods while achieving the lowest TTFT under highly concurrent serving.
Under controlled serving traces, it provides a 2.0$\times$ TTFT speedup over inference without KV cache sharing and reduces peak GPU memory by up to 3.7$\times$ relative to a prior KV cache correction method.
These results establish KVCMAS as an accurate and scalable KV cache sharing approach for prompt-specialized multi-agent serving.
\end{abstract}

\section{Introduction}
\label{sec:intro}

LLM-based multi-agent systems (MAS) improve performance on complex tasks by coordinating agents specialized for complementary roles, such as planning, tool execution, critique, and orchestration
\citep{li2023camel,zhu2025multiagentbench,dong2025insightv,yu2026mact,zhang2024coela,zhang2026mavla,hong2024metagpt,li2025rolediff,li2026morse,zong2024triad}.
Such specialization is implemented by heterogeneous models, role-specific adapters, or distinct role prompts over a shared model
\citep{chen2024reconcile,lee2026mapcoder,kong2024roleplay}.
Among these designs, prompt-based specialization over a shared model is particularly practical for serving because agents share model weights and new roles require no additional training
\citep{kong2024roleplay,wang2024solo,li2025rolediff,wang2026metagen}.
During execution, user queries, retrieved information, tool observations, and agent outputs propagate across agents, forming trajectories that interleave agent-specific prefixes with shared context
\citep{tang2025sdt,li2024massurvey,konstantinova2026a2a}.
As these trajectories grow through multi-turn interactions, additional agents, and multimodal inputs, repeated shared context processing becomes a major serving overhead
\citep{kim2026gaiatrace,zhu2026tracelab,tokendance2026}.

Although prompt-specialized agents share model weights, their different prefixes change the hidden states and KV caches generated for the same shared context
\citep{yao2025cacheblend,liu2026droidspeak,li2026graphflow}.
Consequently, each agent repeatedly prefills the overlapping context and retains a separate KV cache.
Directly reusing a cache constructed under another prefix avoids this redundancy but causes substantial accuracy degradation
\citep{ye2025kvcomm,geng2026relaycaching,ma2026kamera}.
Selective recomputation mitigates this error by rebuilding selected layers or tokens through model execution, but recovering more accuracy requires additional prefill computation
\citep{yao2025cacheblend,liu2026droidspeak,geng2026relaycaching}.
Delta correction instead estimates the cache difference induced by the target context and applies it without executing the model for the reconstruction
\citep{li2026graphflow,ma2026kamera,ye2025kvcomm}.
Most existing correction methods construct relation-specific corrections for recurring context relations
\citep{li2026graphflow,ma2026kamera}, but these corrections remain tied to previously observed context, making accurate correction challenging for new user requests and agent outputs.
KVComm supports dynamically changing context through an online anchor pool, but its full-dimensional base caches and agent-specific corrections cause memory usage to grow with the shared context length
\citep{ye2025kvcomm}.
Existing delta correction methods also construct agent-specific caches from a separate context-free or position-independent reference.
For first seen shared context, constructing this reference requires an additional prefill outside the actual workflow.
Moreover, correction error at the first agent produces an inaccurate output that propagates to subsequent agents as shared context.

To address these limitations, we propose \textbf{KVCMAS}, an online KV cache correction framework for dynamically shared context in prompt-specialized multi-agent systems.
One of our key ideas is that cross-agent KV cache deviations are concentrated in a low-dimensional feature space, allowing KVCMAS to store online correction states in compact low-rank form.
We further observe that using an exact first-agent cache and workflow-relative references reduces aggregate correction error while eliminating a separate context-free reference prefill.
Based on this observation, KVCMAS densely prefills the first agent and uses each corrected shared context cache as the reference for the next workflow edge.
Across text and vision workloads, KVCMAS retains competitive multi-agent task accuracy while reducing correction memory and improving serving efficiency.

\section{Background}
\label{sec:background}

\begin{figure*}[t]
    \centering
    \vspace{-20pt}
    \includegraphics[width=\textwidth]{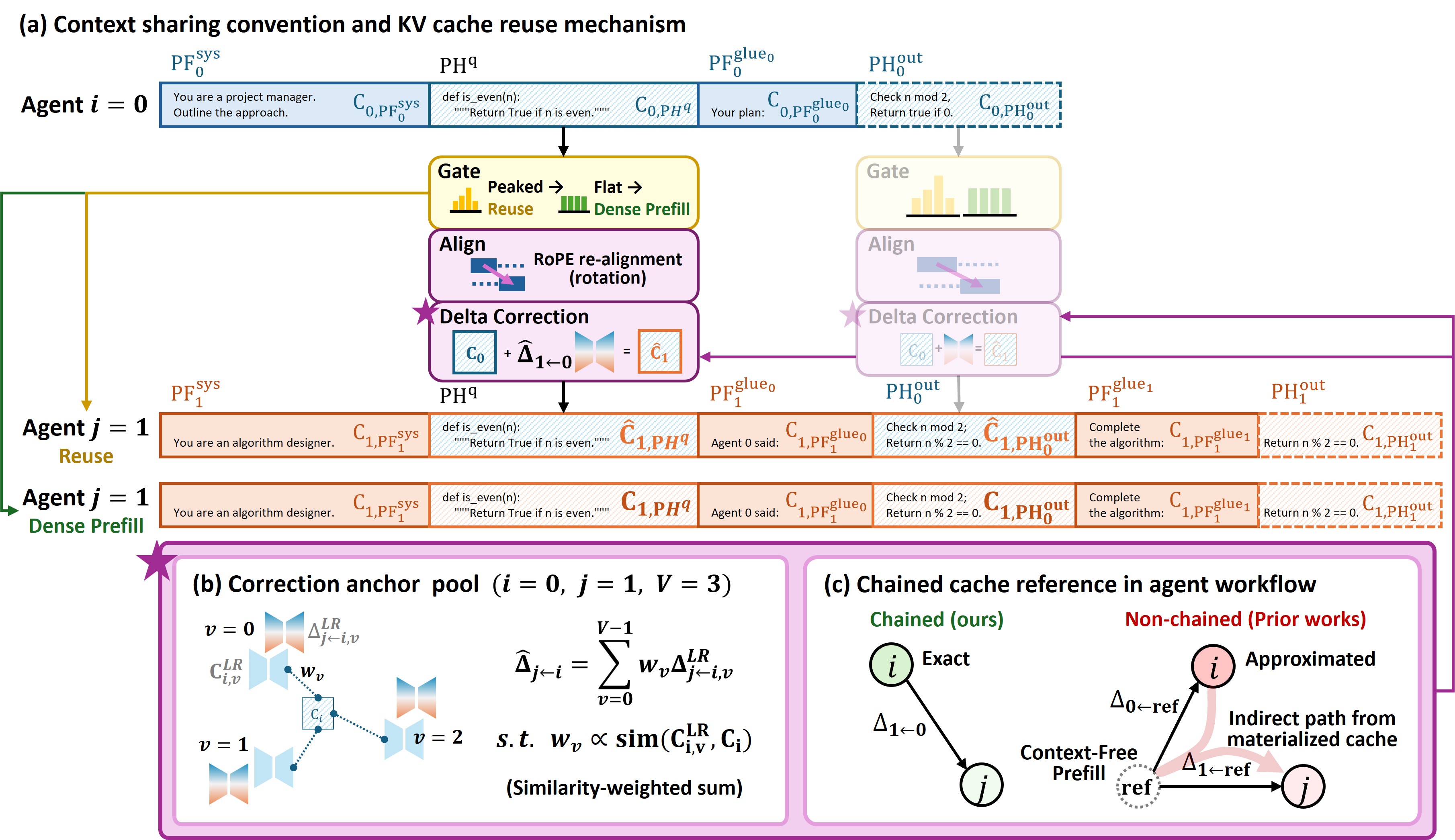}
    \vspace{-15pt}
    \caption{
    Overview of the context sharing convention and KVCMAS on an agent workflow edge from agent $i$ to agent $j$.
    (a) Agent trajectories interleave agent-specific prefix segments (PF) with shared context segments (PH). KVCMAS reuses the preceding agent's KV cache for a PH segment when its correction is reliable and otherwise falls back to dense prefill.
    (b) KVCMAS matches the current cache representation to the online anchors of the corresponding shared segment and estimates the required KV cache correction by combining their low-rank corrections with similarity-based weights.
    (c) The corrected cache becomes the reference for the next workflow edge, chaining corrections along the agent workflow without constructing a separate context-free reference cache.
    }
    \vspace{-5pt}
    \label{fig:overview}
\end{figure*}

\subsection{Context Sharing Convention}
\label{sec:background_sharing}

We consider a MAS in which multiple agents share the same model weights but use different agent-specific prefixes.
As illustrated in Figure~\ref{fig:overview}(a), the trajectory of agent $i$ interleaves agent-specific prefix segments (PF) with shared context segments (PH):
\begin{equation}
x_i =
\big[
\mathrm{PF}^{\mathrm{sys}}_i,\,
\mathrm{PH}^{\mathrm{q}},\,
\mathrm{PF}^{\mathrm{glue}_0}_i,\,
\mathrm{PH}^{\mathrm{out}}_0,\,
\mathrm{PF}^{\mathrm{glue}_1}_i,\,
\mathrm{PH}^{\mathrm{out}}_1,\,
\dots
\big]
\label{eq:pfph}
\end{equation}
Here, $\mathrm{PF}^{\mathrm{sys}}_i$ denotes the system prompt of agent $i$, while $\mathrm{PF}^{\mathrm{glue}_k}_i$ denotes the agent-specific glue prompt placed before the output of agent $k$.
In contrast, $\mathrm{PH}^{\mathrm{q}}$ contains the user query and task observations, while $\mathrm{PH}^{\mathrm{out}}_k$ contains the output of agent $k$.
These PH segments form the shared context propagated across downstream agents.
A downstream agent $j$ follows the same structure with its own PF segments, and its shared context additionally includes $\mathrm{PH}^{\mathrm{out}}_i$ after an edge $i\rightarrow j$.
Such an edge represents an arbitrary agent transition, including those in sequential, loop, fan-in, and fan-out workflows.

For a segment $\mathrm{seg}$ in the trajectory of agent $i$, we denote its preceding context by $c_{i,\mathrm{seg}}$ and its resulting KV cache by $C_{i,\mathrm{seg}}\in\mathbb{R}^{L_{\mathrm{seg}}\times D}$, where $L_{\mathrm{seg}}$ is the segment length and $D$ is the KV cache feature dimension:
\begin{equation}
C_{i,\mathrm{seg}}
=
\mathrm{KV}\!\left(
\mathrm{seg} \mid c_{i,\mathrm{seg}}
\right)
\label{eq:segment_kv}
\end{equation}
Here, $\mathrm{KV}(\mathrm{seg}\mid c)$ denotes the KV cache constructed for segment $\mathrm{seg}$ under preceding context $c$.
An agent-specific PF segment differs in content across agents and is therefore not a target for cross-agent KV cache sharing.
In contrast, a PH segment contains the same content across agents and provides the main opportunity for KV cache sharing, while its preceding context differs because of their agent-specific PF segments.
Consequently, the same PH segment generally produces different KV caches across agents, i.e., $C_{i,\mathrm{seg}} \neq C_{j,\mathrm{seg}}$.
This cross-agent cache deviation prevents direct reuse of otherwise overlapping PH caches.
We next describe how existing KV cache sharing methods address this deviation.

\subsection{KV Cache Sharing Methods}
\label{sec:background_methods}

KV cache sharing reuses a cache previously constructed for an overlapping PH segment, avoiding repeated prefill.
However, the reused cache must account for cross-agent cache deviation to preserve accuracy.
Existing methods address this deviation through selective recomputation or delta correction.
Here, we note that all evaluated methods, including KVCMAS, re-align cached keys to their target positions before cross-agent reuse.
We omit this deterministic alignment from the notation below.

\paragraph{Selective recomputation.}
Selective recomputation replaces a selected subset of the reused KV cache with the corresponding cache recomputed under the target agent's context.
Let $C_{i,\mathrm{seg}}$ be the reused cache, $C_{j,\mathrm{seg}}$ be the target cache, and $\mathcal{T}$ be the cache subset selected for recomputation, which corresponds to layers, tokens, or their combination.
The resulting cache $\widehat{C}_{j,\mathrm{seg}}$ is
\begin{equation}
\widehat{C}_{j,\mathrm{seg}}[u]
=
\begin{cases}
C_{j,\mathrm{seg}}[u], & u \in \mathcal{T}\\
C_{i,\mathrm{seg}}[u], & u \notin \mathcal{T}
\end{cases}
\label{eq:selective_recompute}
\end{equation}
where $u$ denotes a generic cache index.
DroidSpeak identifies critical layer groups by profiling layer-wise KV cache deviation offline and recomputes the full shared context within the selected layers
\citep{liu2026droidspeak}.
CacheBlend instead selects tokens based on KV cache deviation measured in early layers and recomputes the selected tokens through subsequent layers
\citep{yao2025cacheblend}.
RelayCaching further combines KV cache deviation with attention scores to select important tokens and restricts their recomputation to critical middle layers
\citep{geng2026relaycaching}.
However, selective recomputation corrects only the selected cache subset, leaving cache deviations outside $\mathcal{T}$ in the reused cache.
Reducing this remaining deviation requires recomputing a larger subset, which increases prefill computation and produces a trade-off between accuracy and efficiency.
We provide a detailed analysis of the limited reconstruction coverage of selective recomputation in Appendix~\ref{app:obs_recompute}.

\paragraph{Delta correction.}
Delta correction estimates the KV cache deviation from a reference cache to the target agent cache and applies it without executing the model over the corrected entries.
Let $C_{\mathrm{ref},\mathrm{seg}}$ denote the reference cache for segment $\mathrm{seg}$ and $C_{j,\mathrm{seg}}$ denote its target cache for agent $j$.
In the correction methods considered in this work, $C_{\mathrm{ref},\mathrm{seg}}$ is generated from the shared segment without an agent-specific prefix and serves as a context-free reference.
The corrected cache is given by
\begin{equation}
\widehat{C}_{j,\mathrm{seg}}
=
C_{\mathrm{ref},\mathrm{seg}}
+
\widehat{\Delta}_{j\leftarrow\mathrm{ref},\mathrm{seg}}
\label{eq:delta_correction}
\end{equation}
where $\widehat{\Delta}_{j\leftarrow\mathrm{ref},\mathrm{seg}}$ estimates the exact cache deviation $\Delta_{j\leftarrow\mathrm{ref},\mathrm{seg}} = C_{j,\mathrm{seg}} - C_{\mathrm{ref},\mathrm{seg}}$.
When the estimated correction is accepted, delta correction avoids the model execution required by selective recomputation, while an unreliable correction falls back to dense prefill.

GraphFlow constructs and reuses corrections for recurring transitions between agent operations
\citep{li2026graphflow}.
Kamera similarly constructs corrections for repeated multimodal content based on the context that precedes it
\citep{ma2026kamera}.
However, because the task-specific context carried by PH changes across requests and interactions, a correction constructed for a previously observed context relation does not represent new user requests or agent outputs.
Therefore, in dynamic context settings, correction reuse is limited, requiring a newly constructed correction or dense prefill.
KVComm supports dynamically changing shared context using an online anchor pool
\citep{ye2025kvcomm}.
Each anchor stores a context-free base cache and the corresponding agent-specific delta corrections.
For a new segment, KVComm compares its base cache with the stored anchors and estimates the correction as their weighted combination:
\begin{equation}
\widehat{\Delta}_{j\leftarrow\mathrm{ref},\mathrm{seg}}
=
\sum_{v=0}^{V-1}
w_v
\Delta_{j\leftarrow\mathrm{ref},\mathrm{seg}}^{(v)},
\quad
\sum_{v=0}^{V-1} w_v = 1
\label{eq:online_delta}
\end{equation}
where $V$ is the number of anchors and $w_v$ is the normalized weight determined by the similarity between the current and stored base caches.
An unreliable match falls back to dense prefill and provides a new observed correction.
However, storing both the base caches and delta corrections in full-dimensional form makes memory usage grow with the shared context length, the number of anchors, and the number of agents.
Furthermore, the delta correction methods above construct their corrected caches relative to a separately constructed context-free or position-independent reference.
For first seen shared context, constructing this reference adds a prefill outside the agent workflow.
Moreover, correction error at the first agent produces an inaccurate output that propagates to subsequent agents as shared context.
This early error substantially increases the aggregate error across the workflow.
We also note that works using specialized transfer mechanisms for KV cache sharing require additional training or calibration
\citep{cache2cache,heo2026crossmodel,kvlink},
while system-level cache optimizations are complementary to cross-agent cache correction
\citep{kvshare,gim2024promptcache,tokendance2026,zhang2026cachescout,kvflow}.

\section{Methodology}
\label{sec:method}

In this section, we present \textbf{KVCMAS}, an online KV cache correction framework for dynamically shared context in prompt-specialized multi-agent systems.
As illustrated in Figure~\ref{fig:overview}, KVCMAS consists of two major components.
First, it stores the base representations and delta corrections of the online anchor pool in low-rank form, reducing the memory cost of online correction.
Second, it defines each correction relative to the source agent cache already produced along the current workflow edge, avoiding a separately constructed context-free reference cache.
We first present the observations behind these designs and then describe the KVCMAS correction procedure.
Unless otherwise noted, the observations use the workloads and models described in Section~\ref{sec:experiments}.

\subsection{Compact Online Correction via Low-Rank Anchor Pools}
\label{sec:method_lr}

\begin{figure*}[t]
    \centering
    \vspace{-15pt}
    \includegraphics[width=\textwidth]{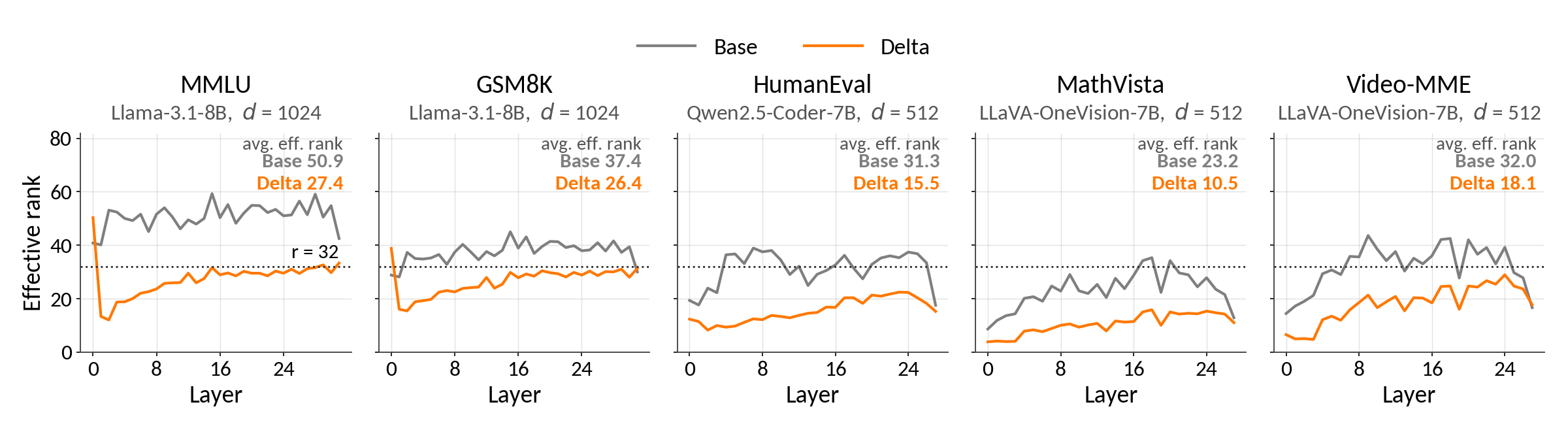}
    \vspace{-25pt}
    \caption{
    Effective rank of the base cache and delta correction across layers and workloads.
    Effective rank is defined as the minimum rank retaining 90\% of the singular-value energy.
    The annotated values report the effective rank averaged across layers, and the dotted line marks rank 32.
    }
    \label{fig:obs_rank}
\end{figure*}

Online delta correction incurs substantial memory overhead because each anchor stores a full-dimensional base cache and agent-specific correction states.
Our key observation is that these online correction states occupy a low-dimensional feature space.
KVCMAS exploits this structure to compress the anchor pool while leaving the active KV cache used by attention full-dimensional.
For a shared context of length $L$ and KV cache feature dimension $D$, storing these states across $V$ anchors and $N$ consuming agents requires $\mathcal{O}(VNLD)$ memory for each pool.
For example, with Llama-3.1-8B in BF16, a 4K-token segment pool with $V=20$ and corrections for $N=4$ agents requires approximately 50\,GiB for the full-dimensional anchor states alone.

Specifically, we observe that agent-specific delta corrections have an effective rank below 32 across the various workloads, substantially below the full feature dimension (Figure~\ref{fig:obs_rank}).
The base caches exhibit a similar structure, although their effective ranks are slightly higher, consistent with prior observations on KV caches
\citep{chang2025palu,saxena2024eigen}.
KVCMAS therefore stores both base caches and delta corrections in low-rank form.
The reconstructed base representations are used only for anchor matching, while the reconstructed correction is applied to the source-agent cache to materialize a full-dimensional target cache.
Appendix~\ref{app:obs_pf} further shows that PF-induced deviations have substantially lower effective rank than PH-induced deviations, highlighting the greater challenge of correcting dynamically changing PH.

\subsection{Chained Correction along the Agent Workflow}
\label{sec:method_chain}

\begin{figure*}[t]
    \centering
    \vspace{-10pt}
    \includegraphics[width=\textwidth]{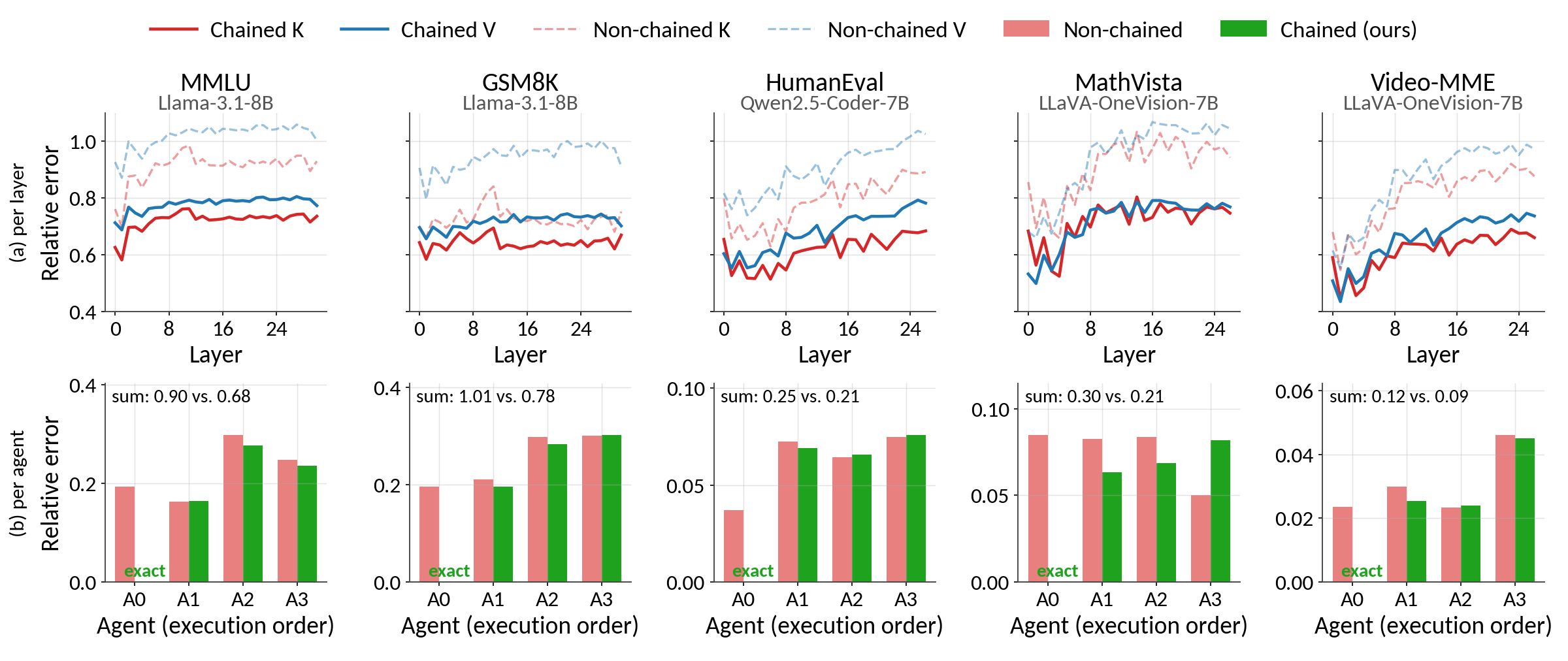}
    \vspace{-20pt}
    \caption{
    KV cache reconstruction error normalized by the Frobenius norm of the target tensor, using a context-free reference (non-chained) and the source agent cache (chained).
    The error is measured against the exact cache obtained through full prefill.
    (a) reports the error across layers, and (b) reports the error across agents in execution order A0$\rightarrow$A1$\rightarrow$A2$\rightarrow$A3, with an exact A0 cache under chained correction.
    }
    \label{fig:obs_chain}
\end{figure*}

Existing delta correction methods construct each agent-specific cache relative to a context-free reference $C_{\mathrm{ref},\mathrm{seg}}$.
We refer to this reference structure as \textit{non-chained} correction.
Because $C_{\mathrm{ref},\mathrm{seg}}$ is not produced by an executed agent, constructing it for first seen shared context requires a separate prefill outside the actual workflow.
Moreover, when only the materialized agent cache $\widehat{C}_{i,\mathrm{seg}}$ is carried forward, transitioning to agent $j$ requires either retaining $C_{\mathrm{ref},\mathrm{seg}}$ separately or recovering it before applying
$\widehat{\Delta}_{j\leftarrow\mathrm{ref},\mathrm{seg}}$.
This produces the additional reference construction and indirect correction path illustrated in Figure~\ref{fig:overview}(c).
Non-chained correction also applies an approximate correction to the first agent, and the resulting cache error produces an inaccurate output that propagates to subsequent agents as another shared context.

In contrast, the source agent cache $C_{i,\mathrm{seg}}$ is already produced by the actual workflow and reflects the preceding interaction trajectory.
The exact correction from source agent $i$ to target agent $j$ is
\begin{equation}
\Delta_{j\leftarrow i,\mathrm{seg}}
=
C_{j,\mathrm{seg}}-C_{i,\mathrm{seg}}
=
\Delta_{j\leftarrow\mathrm{ref},\mathrm{seg}}
-
\Delta_{i\leftarrow\mathrm{ref},\mathrm{seg}}
\label{eq:chained_delta}
\end{equation}
We refer to correction relative to $C_{i,\mathrm{seg}}$ as \textit{chained} correction because its reference follows the agent workflow.
With exact corrections, the context-free and source agent references are algebraically equivalent.
However, with empirical corrections obtained from actual model runs, the selected reference produces different approximation errors.
KVCMAS densely prefills the first agent and chains correction from each materialized shared context cache.
As shown in Figure~\ref{fig:obs_chain}, this structure reduces aggregate correction error across the evaluated layers and agents.

\subsection{KVCMAS}
\label{sec:method_kvcmas}

Building on the two core ideas introduced above, KVCMAS combines low-rank online correction with chained correction along the agent workflow, as illustrated in Figure~\ref{fig:overview}.
KVCMAS processes agent-specific PF separately and applies online correction to shared segments.
As shown in Figure~\ref{fig:overview}(a), the first agent densely prefills its input and produces an exact cache.
For a shared segment transferred from source agent $i$ to target agent $j$, KVCMAS re-aligns the key positions in $C_{i,\mathrm{seg}}$, matches the source cache to the online anchors, and estimates the required correction.
Since $C_{i,\mathrm{seg}}$ is already produced by the source agent, anchor matching requires no additional model forward pass.

KVCMAS maintains an anchor pool for each shared placeholder slot.
Each anchor stores a base cache representation together with corrections for the consuming agents observed for that slot.
Figure~\ref{fig:overview}(b) shows the low-rank states stored in each anchor.
For compact notation, let
$Z\in\{C_{i,\mathrm{seg}}^{(v)},\Delta_{j\leftarrow i,\mathrm{seg}}^{(v)}\}$
denote either the base cache or delta correction stored in anchor $v$.
For each key or value matrix
$Z\in\mathbb{R}^{L_{\mathrm{seg}}\times D}$, KVCMAS applies truncated SVD:
\begin{equation}
Z
=
U_Z\Sigma_ZR_Z^\top
\approx
U_{Z,r}\Sigma_{Z,r}R_{Z,r}^\top
=
A_ZB_Z,
\quad
A_Z
=
U_{Z,r}\Sigma_{Z,r},
\;\;
B_Z
=
R_{Z,r}^\top
\label{eq:low_rank_state}
\end{equation}
where the subscript $r$ denotes the components corresponding to the largest $r$ singular values,
$A_Z\in\mathbb{R}^{L_{\mathrm{seg}}\times r}$, and
$B_Z\in\mathbb{R}^{r\times D}$.
Each anchor $v$ stores the low-rank factors of its base cache
$C_{i,\mathrm{seg}}^{(v)}$ and the available corrections
$\Delta_{j\leftarrow i,\mathrm{seg}}^{(v)}$ for its consuming agents.
KVCMAS compares $C_{i,\mathrm{seg}}$ with the reconstructed base representation of each anchor and assigns larger weights to more similar anchors.
Following Eq.~\ref{eq:online_delta}, it estimates the transition correction directly from the stored low-rank factors:
\begin{equation}
\widehat{\Delta}_{j\leftarrow i,\mathrm{seg}}
=
\sum_{v=0}^{V-1}
w_v
A_{\Delta,v}B_{\Delta,v},
\quad
\sum_{v=0}^{V-1}w_v=1
\label{eq:kvcmas_delta}
\end{equation}
where $A_{\Delta,v}B_{\Delta,v}$ represents the delta correction stored in anchor $v$ and $w_v$ denotes its normalized similarity weight.
For a pool containing corrections for $N$ consuming agents, this reduces its anchor memory from
$\mathcal{O}(VNLD)$ to $\mathcal{O}(VNr(L+D))$.

Following KVComm~\citep{ye2025kvcomm}, KVCMAS determines correction reliability from the normalized entropy of the anchor matching scores.
A peaked distribution indicates a distinctive anchor match, whereas a flat distribution triggers dense prefill.
A correction is accepted when its normalized entropy does not exceed the threshold $\tau$, with a higher $\tau$ yielding a higher reuse ratio $\rho$.
For an accepted correction, KVCMAS applies
$\widehat{\Delta}_{j\leftarrow i,\mathrm{seg}}$
to the aligned source agent cache and materializes the full-dimensional target cache
$\widehat{C}_{j,\mathrm{seg}}$.
This cache is used by ordinary attention and becomes the shared context reference for the next workflow edge, realizing the chained correction in Eq.~\ref{eq:chained_delta} without returning to a context-free reference cache.
Otherwise, agent $j$ densely prefills the segment and produces an exact cache.
KVCMAS factorizes the current source cache and its observed transition correction when updating the corresponding anchor pool.
Further details on the anchor pool management are provided in Appendix~\ref{app:obs_anchor}.

\section{Experiments}
\label{sec:experiments}

\subsection{Experimental Setup}
\label{sec:exp_setup}

\textbf{Workloads and Baselines.}
We evaluate KVCMAS using the multi-agent framework of KVComm~\citep{ye2025kvcomm}, which builds on AgentPrune~\citep{zhang2025agentprune} and GPTSwarm~\citep{zhuge2024gptswarm}.
We use Llama-3.1-8B-Instruct for MMLU and GSM8K, Qwen2.5-Coder-7B-Instruct for HumanEval, and LLaVA-OneVision-7B for MathVista and Video-MME.
The LLM workloads follow the agent prompts of KVComm, while MathVista and Video-MME adapt the prompt structures of GSM8K and MMLU, respectively.
Each accuracy workload uses three task-specific agents followed by a reflection agent.
We include execution without KV cache sharing, which repeatedly prefills the shared context (\textbf{NonShared}), and direct cross-agent cache reuse without correction (\textbf{FullShared}) as reference baselines, together with DroidSpeak~\citep{liu2026droidspeak}, CacheBlend~\citep{yao2025cacheblend}, RelayCaching~\citep{geng2026relaycaching}, GraphFlow~\citep{li2026graphflow}, and KVComm~\citep{ye2025kvcomm}.
Since Kamera~\citep{ma2026kamera} similarly relies on corrections constructed for previously observed context relations, we use GraphFlow as the representative baseline for this setting; its results demonstrate the limitation of preconstructed corrections for dynamically changing shared context.
Detailed workload and baseline configurations are provided in Appendix~\ref{app:exp_setup}.

\textbf{Accuracy Evaluation.}
We denote $\rho$ by the fraction of shared cache reused, corresponding to entries excluded from selective recomputation or served through accepted delta correction.
Equal $\rho$ does not imply equal computation because selective recomputation executes the model over recomputed entries, whereas delta correction directly updates accepted entries~\cite{ye2025kvcomm, ma2026kamera}.
We evaluate selective recomputation at $\rho=0.9$ and $0.8$ and delta correction at approximately $\rho=0.8$ and $0.6$ for the LLM and VLM workloads, respectively.
For methods with online anchor matching, we select the reliability threshold $\tau$ to reach these operating points.
Unless otherwise stated, KVCMAS uses rank $r=32$ for both base caches and delta corrections and maintains up to $V=10$ anchors per shared context slot.
Accuracy is averaged over three runs, while latency, TTFT, and peak memory are measured on an NVIDIA A100 80\,GB GPU at 1 QPS, with TTFT averaged across each trajectory.

\textbf{Serving Efficiency.}
We implement the methods upon vLLM~\citep{kwon2023vllm} and evaluate them using controlled agent traces across a QPS sweep.
The traces fix a four-agent schedule and incorporate 128-token agent outputs into subsequent inputs while varying the non-generated shared context.
We report request-level median (p50) and tail (p90) TTFT.
We compare selective recomputation at $\rho=0.9$ and $0.8$ with delta correction at $\rho=0.8$ and $0.6$, respectively.
These concurrent serving experiments run on an NVIDIA A100 80\,GB GPU.
To examine per-request single-batched efficiency, we measure memory usage and throughput under single-stream execution on an NVIDIA A6000 48\,GB GPU.
Using the same controlled setting, we further vary the rank, anchor pool size, number of agents, and number of interaction rounds in Appendix~\ref{app:abl}.

\subsection{Benchmark Accuracy}
\label{sec:exp_acc}

\begin{table*}[t]
\centering
\vspace{-15pt}
\caption{
Accuracy (Acc., \%), end-to-end latency (Lat., s), TTFT (s), peak GPU memory (Mem., GB), and KV cache reuse ratio ($\rho$).
Excluding NonShared, the highest and second-highest accuracy among KV cache sharing methods are shown in \textbf{bold} and \underline{underlined}, respectively.
}
\label{tab:acc}
\setlength{\tabcolsep}{2.8pt}

\textbf{(a) LLM benchmarks}\\[2pt]
\resizebox{\textwidth}{!}{
\begin{tabular}{lccccccccccccccc}
\toprule
& \multicolumn{5}{c}{\textbf{MMLU}} & \multicolumn{5}{c}{\textbf{GSM8K}} & \multicolumn{5}{c}{\textbf{HumanEval}} \\
\cmidrule(lr){2-6}\cmidrule(lr){7-11}\cmidrule(lr){12-16}
\textbf{Method} & \textbf{Acc.$\uparrow$} & \textbf{Lat.$\downarrow$} & \textbf{\footnotesize TTFT$\downarrow$} & \textbf{Mem.$\downarrow$} & $\boldsymbol{\rho}$ & \textbf{Acc.$\uparrow$} & \textbf{Lat.$\downarrow$} & \textbf{\footnotesize TTFT$\downarrow$} & \textbf{Mem.$\downarrow$} & $\boldsymbol{\rho}$ & \textbf{Pass@1$\uparrow$} & \textbf{Lat.$\downarrow$} & \textbf{\footnotesize TTFT$\downarrow$} & \textbf{Mem.$\downarrow$} & $\boldsymbol{\rho}$ \\
\midrule
NonShared & 67.54 & 18.59 & 0.06 & 16.78 & 0.00 & 85.60 & 20.82 & 0.12 & 16.88 & 0.00 & 86.96 & 8.80 & 0.05 & 15.28 & 0.00 \\
FullShared & 36.45 & 10.52 & 0.04 & 15.92 & 1.00 & 29.42 & 23.87 & 0.12 & 16.14 & 1.00 & 80.75 & 7.90 & 0.04 & 15.04 & 1.00 \\
DroidSpeak & 41.61 & 17.10 & 0.06 & 16.07 & 0.91 & 37.83 & 20.50 & 0.13 & 16.30 & 0.91 & 82.61 & 8.84 & 0.05 & 15.08 & 0.89 \\
CacheBlend & 55.39 & 18.49 & 0.07 & 16.15 & 0.90 & 77.03 & 19.49 & 0.12 & 16.29 & 0.90 & 82.61 & 8.34 & 0.05 & 15.02 & 0.90 \\
RelayCaching & 60.68 & 17.76 & 0.05 & 16.23 & 0.90 & 77.94 & 20.75 & 0.12 & 16.27 & 0.90 & 83.23 & 8.45 & 0.05 & 15.01 & 0.90 \\
GraphFlow & 36.38 & 21.36 & 0.09 & 16.12 & 0.82 & 31.46 & 30.17 & 0.18 & 16.46 & 0.79 & 60.87 & 8.82 & 0.05 & 15.10 & 0.83 \\
KVComm & \underline{62.57} & 15.37 & 0.08 & 52.40 & 0.76 & \textbf{80.14} & 21.49 & 0.16 & 59.18 & 0.80 & \textbf{85.09} & 8.55 & 0.05 & 30.54 & 0.80 \\
\textbf{KVCMAS} & \textbf{65.45} & 14.21 & 0.05 & 16.06 & 0.79 & \underline{79.53} & 19.71 & 0.12 & 16.10 & 0.76 & \underline{84.47} & 8.32 & 0.05 & 15.04 & 0.78 \\
\bottomrule
\end{tabular}
}

\vspace{5pt}
\textbf{(b) VLM benchmarks}\\[2pt]
\resizebox{0.72\textwidth}{!}{
\begin{tabular}{lcccccccccc}
\toprule
& \multicolumn{5}{c}{\textbf{MathVista}} & \multicolumn{5}{c}{\textbf{Video-MME}} \\
\cmidrule(lr){2-6}\cmidrule(lr){7-11}
\textbf{Method} & \textbf{Acc.$\uparrow$} & \textbf{Lat.$\downarrow$} & \textbf{\footnotesize TTFT$\downarrow$} & \textbf{Mem.$\downarrow$} & $\boldsymbol{\rho}$ & \textbf{Acc.$\uparrow$} & \textbf{Lat.$\downarrow$} & \textbf{\footnotesize TTFT$\downarrow$} & \textbf{Mem.$\downarrow$} & $\boldsymbol{\rho}$ \\
\midrule
NonShared & 55.90 & 4.56 & 0.55 & 17.43 & 0.00 & 53.00 & 183.66 & 37.76 & 17.60 & 0.00 \\
FullShared & 50.20 & 0.50 & 0.26 & 16.70 & 1.00 & 37.25 & 22.65 & 0.48 & 16.74 & 1.00 \\
DroidSpeak & 53.00 & 3.49 & 0.46 & 16.86 & 0.79 & 48.84 & 311.84 & 29.53 & 16.88 & 0.79 \\
CacheBlend & 50.80 & 3.98 & 0.49 & 16.86 & 0.80 & 46.36 & 237.25 & 38.76 & 16.89 & 0.80 \\
RelayCaching & 50.50 & 3.96 & 0.49 & 16.89 & 0.80 & 44.92 & 243.64 & 40.72 & 16.86 & 0.80 \\
GraphFlow & 34.60 & 2.30 & 0.41 & 16.87 & 0.64 & 28.22 & 190.46 & 25.76 & 16.64 & 0.62 \\
KVComm & \underline{53.50} & 2.43 & 0.42 & 59.19 & 0.63 & \underline{51.56} & 163.06 & 24.30 & 53.27 & 0.62 \\
\textbf{KVCMAS} & \textbf{53.80} & 1.72 & 0.34 & 16.77 & 0.63 & \textbf{52.36} & 151.67 & 24.06 & 16.61 & 0.64 \\
\bottomrule
\end{tabular}
}
\vspace{-5pt}
\end{table*}

\begin{figure*}[t]
    \centering
    \vspace{-10pt}
    \includegraphics[width=\textwidth]{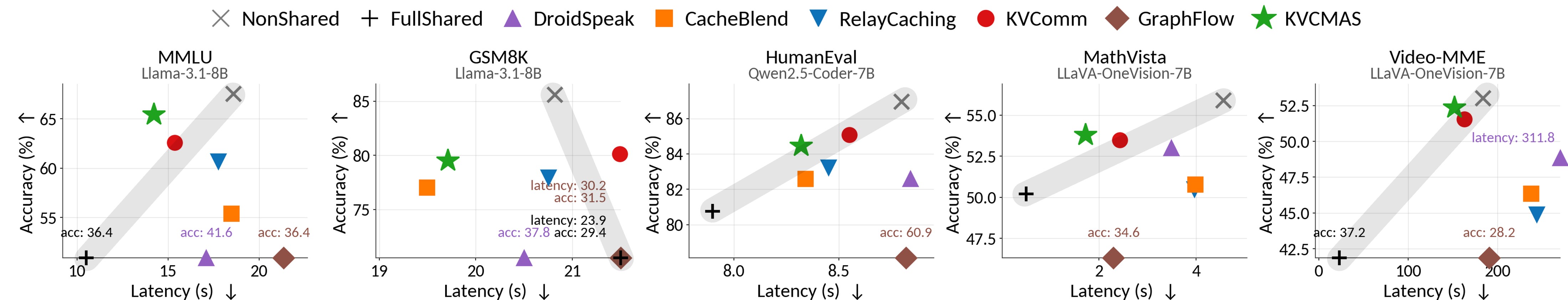}
    \vspace{-20pt}
\caption{
Benchmark accuracy and latency trade-off across the evaluated workloads.
The shaded region spans FullShared and NonShared, with the upper left indicating a better trade-off.
Out-of-range results are placed on the plot boundaries and annotated with their values.
KVCMAS consistently remains near the upper-left Pareto frontier.
}
    \label{fig:pareto}
\end{figure*}

Table~\ref{tab:acc} reports accuracy and system performance across all workloads.
KVCMAS achieves the highest accuracy among KV cache sharing methods on three of the five benchmarks and ranks second to KVComm on GSM8K and HumanEval within $0.7\%$ points.
It also provides the lowest end-to-end latency and TTFT among the delta correction methods across all workloads and achieves lower end-to-end latency than the selective recomputation methods except on GSM8K, despite using lower reuse ratios.
Consequently, KVCMAS remains near the upper-left Pareto frontier in Figure~\ref{fig:pareto}, providing the strongest overall accuracy and latency trade-off.

Selective recomputation partially recovers the accuracy loss of FullShared by rebuilding a selected cache subset.
However, deviations outside the selected subset remain uncorrected, while increasing its coverage requires additional model execution.
This trade-off is particularly evident on Video-MME and aligns with the reconstruction analysis in Appendix~\ref{app:obs_recompute}.
Delta correction instead updates the reused cache without model execution.
GraphFlow applies corrections constructed for previously observed context relations without online matching to the current shared context, leading to substantial accuracy degradation on dynamic requests.
KVComm recovers much of this loss through online anchor matching, but its full-dimensional anchor pool uses up to 3.7$\times$ more peak GPU memory than KVCMAS.
KVCMAS retains online matching while preserving an exact first-agent cache, chaining corrections along the workflow, and storing anchor states in low-rank form.
It provides accuracy comparable to or higher than that of KVComm, with lower latency and memory usage.
Because benchmark latency also reflects differences in generated responses and agent trajectories, Section~\ref{sec:exp_eff} evaluates serving efficiency using controlled traces.
In addition, Appendices~\ref{app:exp_acc} and~\ref{app:exp_th} report run-to-run variation, trajectory lengths, and ablations of the reliability threshold.

\subsection{Serving Efficiency}
\label{sec:exp_eff}

\begin{figure*}[t]
    \centering
    \includegraphics[width=\textwidth]{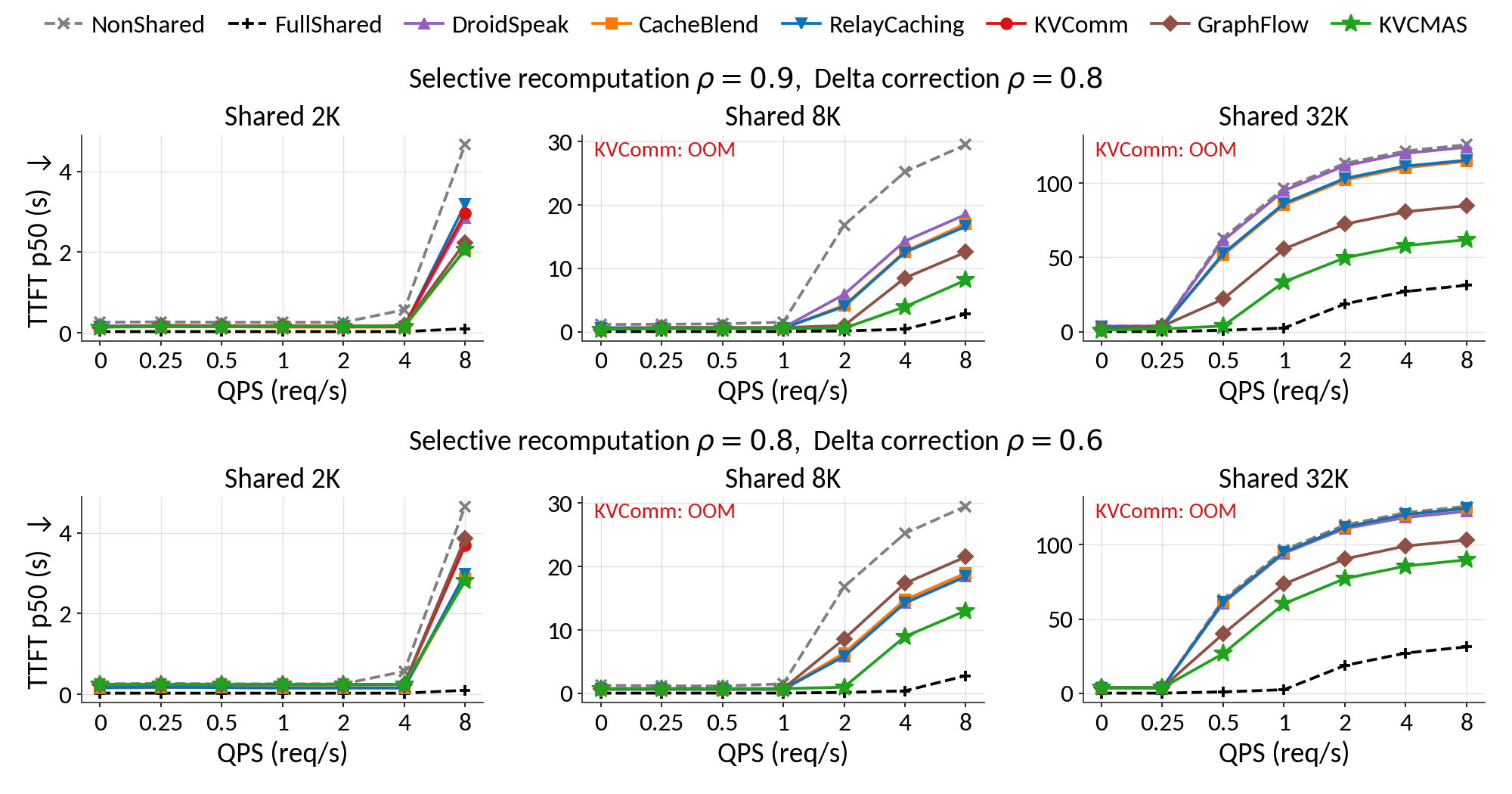}
    \vspace{-25pt}
    \caption{
    Median (p50) TTFT under concurrent serving across shared context lengths and QPS.
    The top row compares selective recomputation at $\rho=0.9$ with delta correction at $\rho=0.8$, while the bottom row compares $\rho=0.8$ with $\rho=0.6$, respectively.
    KVComm runs out of memory at 8K and 32K shared context.
    }
    \label{fig:serve_eff}
    \vspace{-5pt}
\end{figure*}

\begin{figure*}[t]
    \centering
    \vspace{-10pt}
    \includegraphics[width=\textwidth]{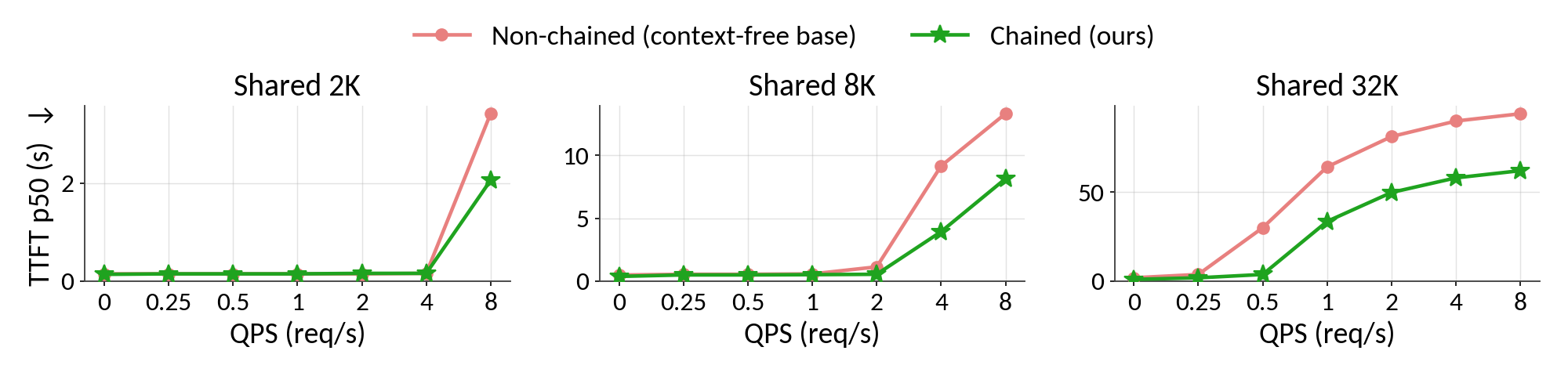}
    \vspace{-25pt}
    \caption{
    Median (p50) TTFT of chained and non-chained correction at $\rho=0.8$ across shared context lengths and QPS.
    Non-chained correction uses a context-free reference, while chained correction follows the preceding agent cache.
    }
    \label{fig:chain_eff}
    \vspace{-5pt}
\end{figure*}

KVCMAS achieves the lowest median TTFT among the KV cache sharing works, with larger advantages at longer shared contexts and higher request rates, as shown in Figure~\ref{fig:serve_eff}.
At 32K shared tokens and 8 QPS under the $\rho=0.8$ setting, KVCMAS provides a 2.0$\times$ TTFT speedup over NonShared and 27\% lower TTFT than GraphFlow, the next-fastest KV cache sharing work.
NonShared repeatedly prefills the shared context, while FullShared avoids this computation through direct reuse at the cost of substantial accuracy degradation.
DroidSpeak propagates hidden states for all shared context through the selected layers to update their KV projections, whereas CacheBlend and RelayCaching restrict model execution to selected tokens.
Delta correction instead updates corrected cache entries without model execution, resulting in lower TTFT for GraphFlow, KVComm, and KVCMAS.

The TTFT gain of KVCMAS compared to previous delta correction methods becomes more pronounced as the shared context grows.
For first seen shared context, non-chained correction constructs a separate context-free reference outside the agent workflow.
KVComm additionally stores its online anchor states in full-dimensional form and runs out of memory at 8K and 32K shared contexts.
KVCMAS stores these states in low-rank form and chains correction from a cache already produced by the workflow, avoiding the reference prefill and reducing anchor memory.
The results at $\rho=0.6$ follow the same scaling trend as those at $\rho=0.8$.
At shorter contexts and lower request rates, where prefill contributes less to serving time, the differences among methods are smaller.

Figure~\ref{fig:chain_eff} compares chained correction with a non-chained variant while keeping the remaining procedure fixed.
At 32K shared tokens and 8 QPS, chained correction reduces median TTFT by 34\% by removing the reference prefill that would otherwise compete with concurrent requests for batching and compute resources.
Appendices~\ref{app:exp_serve} and~\ref{app:exp_chaineff} provide the complete p50 values and additional p90 results for Figures~\ref{fig:serve_eff} and~\ref{fig:chain_eff}, respectively.
Appendix~\ref{app:exp_dec} extends the evaluation to shared context accumulated from agent outputs, where KVCMAS retains its serving efficiency advantage.

Under single-stream execution, KVCMAS uses substantially less memory than KVComm while maintaining peak memory comparable to other methods, and its TTFT and throughput gains increase with shared-context length (Appendix~\ref{app:abl_static}).
The rank and anchor pool size ablations show that $r=32$ and $V=10$ capture most of the accuracy gains, while larger settings provide only marginal improvements with higher memory usage (Appendices~\ref{app:abl_rank} and~\ref{app:abl_anchor}).
KVCMAS also retains the lowest TTFT and highest throughput among the KV cache sharing works as the number of agents and interaction rounds increases (Appendices~\ref{app:abl_agentn} and~\ref{app:abl_round}).

\section{Conclusion}
\label{sec:conclusion}

We introduced KVCMAS, an online KV cache correction framework for dynamically changing shared context in prompt-specialized multi-agent systems.
KVCMAS stores cross-agent cache deviations in compact low-rank form and chains corrections using caches naturally produced along the workflow.
This provides an exact first agent cache, improves correction accuracy, and removes the reference prefill required by non-chained correction.
Across language and vision-language workloads, KVCMAS achieves the highest overall accuracy and the lowest serving latency among KV cache sharing works at high request rates.
At 32K shared tokens and 8 QPS, it provides a 2.0$\times$ TTFT speedup over NonShared, while chained correction reduces TTFT by 34\% compared with non-chained correction.
KVCMAS also reduces peak GPU memory by up to 3.7$\times$ relative to KVComm and retains the lowest TTFT and highest throughput as the number of agents and interaction rounds increases.
These results establish KVCMAS as an accurate and scalable KV cache sharing approach for multi-agent serving.

\newpage






\bibliography{iclr2027_conference}
\bibliographystyle{iclr2027_conference}

\newpage
\appendix

\section{Analysis of KV Cache Reuse}

\subsection{Reconstruction Coverage of Selective Recomputation}
\label{app:obs_recompute}

Section~\ref{sec:exp_acc} shows that selective recomputation incurs accuracy degradation relative to dense prefill (NonShared).
To examine this difference, we measure how much KV cache reuse error is removed by representative layer and token selection criteria.
We use the operating points from the accuracy evaluation: $\rho=0.9$ for the LLM workloads and $\rho=0.8$ for the VLM workloads. Throughout this work, KV cache reuse error is measured as the Frobenius norm of the cache difference normalized by the Frobenius norm of the corresponding full target tensor.

We analyze three criteria used as core components of existing selective recomputation methods.
Layer selection recomputes layers with large KV cache deviation, following the offline profiling criterion of DroidSpeak~\citep{liu2026droidspeak}.
Deviation selection chooses tokens with large KV cache deviation, as used by CacheBlend~\citep{yao2025cacheblend}.
Attention selection chooses tokens with high attention importance, one of the criteria used by RelayCaching~\citep{geng2026relaycaching}.
We note that this analysis isolates each selection criterion rather than reproducing the complete execution procedure of each method.
For each criterion, \textit{Coverage} denotes the fraction of the total KV cache reuse error removed by recomputation.
\textit{Relative Coverage} normalizes this value by random selection under the same recomputation budget.

\begin{table*}[h]
\centering
\caption{
Reconstruction coverage of selective recomputation across datasets and models.
Coverage measures the fraction of KV cache reuse error removed by recomputation, and Relative Coverage normalizes it by random selection under the same budget.
Higher is better for both.
}
\label{tab:obs_recompute}
\resizebox{\textwidth}{!}{
\begin{tabular}{lcccccc}
\toprule
& \multicolumn{2}{c}{\textbf{Layer Selection}}
& \multicolumn{2}{c}{\textbf{Deviation Selection}}
& \multicolumn{2}{c}{\textbf{Attention Selection}} \\
\cmidrule(lr){2-3}
\cmidrule(lr){4-5}
\cmidrule(lr){6-7}
\textbf{Dataset / Model}
& \textbf{Coverage} & \textbf{Relative Coverage}
& \textbf{Coverage} & \textbf{Relative Coverage}
& \textbf{Coverage} & \textbf{Relative Coverage} \\
\midrule
\begin{tabular}[c]{@{}l@{}}GSM8K \\ \scriptsize Llama-3.1-8B\end{tabular} & 0.13 & 1.33$\times$ & 0.16 & 1.61$\times$ & 0.15 & 1.45$\times$ \\
\begin{tabular}[c]{@{}l@{}}MMLU \\ \scriptsize Llama-3.1-8B\end{tabular} & 0.03 & 0.87$\times$ & 0.27 & 2.66$\times$ & 0.17 & 1.65$\times$ \\
\begin{tabular}[c]{@{}l@{}}HumanEval \\ \scriptsize Qwen2.5-Coder-7B\end{tabular} & 0.29 & 2.67$\times$ & 0.32 & 3.24$\times$ & 0.13 & 1.33$\times$ \\
\begin{tabular}[c]{@{}l@{}}MathVista \\ \scriptsize LLaVA-OneVision-7B\end{tabular} & 0.50 & 2.35$\times$ & 0.22 & 1.10$\times$ & 0.20 & 1.02$\times$ \\
\begin{tabular}[c]{@{}l@{}}Video-MME \\ \scriptsize LLaVA-OneVision-7B\end{tabular} & 0.39 & 1.82$\times$ & 0.21 & 1.03$\times$ & 0.20 & 0.99$\times$ \\
\bottomrule
\end{tabular}
}
\end{table*}

Table~\ref{tab:obs_recompute} shows that the evaluated criteria generally identify more informative cache subsets than random selection.
However, at least half of the reuse error remains outside the recomputed subset across all workloads.
On MathVista and Video-MME, deviation and attention selection remove only about 20\% of the error and remain close to random selection.
This limited coverage is consistent with the larger accuracy gap of token selective recomputation on the VLM workloads in Section~\ref{sec:exp_acc}.
Although coverage does not directly determine task accuracy because layers and token positions affect model outputs differently, these results show the difficulty of correcting cross-agent cache deviation under a limited recomputation budget.
KVCMAS instead estimates corrections for the full reused cache without model execution over individual entries.
This broader correction coverage leads to its higher accuracy with lower latency in long-context, high-load regimes despite its lower reuse ratio.

\newpage
\subsection{Low-Rank Structure of PF-Induced Delta}
\label{app:obs_pf}

Section~\ref{sec:method_lr} shows that cross-agent corrections for dynamically shared PH admit compact low-rank representations.
We further examine the cache deviations induced when recurring PF segments, including system and glue prompts, precede a shared segment.
This analysis concerns corrections to the shared cache under different PF contexts rather than reuse of the agent-specific PF caches themselves.
Before computing each delta, we re-align the key positions to remove deterministic differences introduced by RoPE.
Unlike PH, which changes across requests and grows along the workflow, the PF relation for a given agent recurs across requests.
KVCMAS therefore also applies to recurring PF relations: following the preconstructed correction setting of GraphFlow and Kamera~\citep{li2026graphflow,ma2026kamera}, the corresponding delta is profiled once, stored in low-rank form, and positionally re-aligned when reused.

\begin{figure*}[h]
    \centering
    \includegraphics[width=\textwidth]{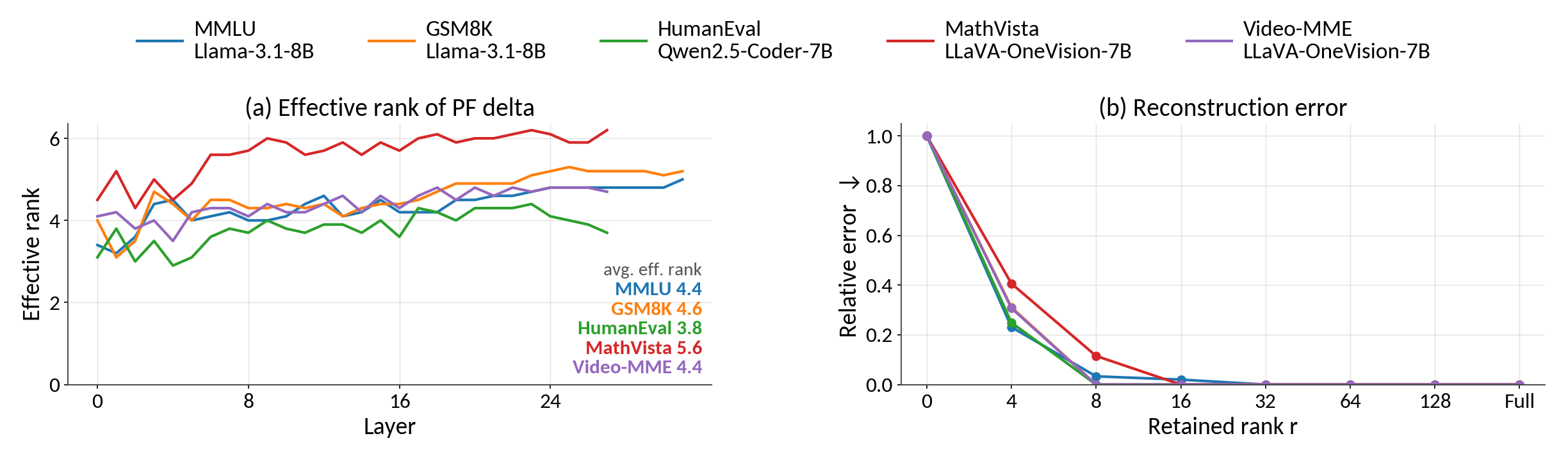}
    \caption{
    Effective rank and reconstruction error of PF-induced delta corrections.
    (a) reports the layer-wise effective rank, defined as the minimum rank retaining 90\% of the singular-value energy, together with the average across layers for each workload.
    (b) reports the relative reconstruction error as the retained rank increases.
    }
    \label{fig:obs_pf}
\end{figure*}

As shown in Figure~\ref{fig:obs_pf}, PF-induced cache deviations exhibit a more compact low-rank structure than deviations from dynamically changing PH.
Their effective rank remains around 4--6 across the evaluated workloads, and rank 8 retains over 90\% of the singular-value energy.
We note that the lengths of both PF and PH segments are greater than their effective ranks, which reflects their actual low-rank structure.
These results support applying the low-rank correction of KVCMAS to recurring PF without online anchor matching.
This work focuses on the challenging dynamically changing PH, for which KVCMAS estimates the required correction online through the anchor pool.

\newpage
\subsection{Anchor Pool Management}
\label{app:obs_anchor}

Section~\ref{sec:method_kvcmas} presents the low-rank online correction used by KVCMAS.
Here, we detail its anchor matching, reliability control, and pool updates.
Each shared context slot maintains a process-wide pool of up to $V$ anchors, each storing low-rank source cache representations and corrections for its consuming workflow nodes.
Following KVComm, KVCMAS uses normalized entropy as a confidence score.
Low entropy indicates that one compatible anchor is distinctly preferred, whereas high entropy indicates that no candidate is clearly preferred.

\paragraph{Reliability Gate.}
For a comparison prefix of length $k$, let $\mathcal{A}_k$ contain the anchors whose stored length is at least $k$.
KVCMAS reconstructs their value states layer by layer and computes:
\begin{equation}
d_v^{\mathrm{gate}}
=
\left(
\sum_{\ell=1}^{N_{\mathrm{layer}}}
\left\|
V_{i,\mathrm{seg}}^{(\ell)}[:k]
-
\widetilde{V}_{i,\mathrm{seg}}^{(v,\ell)}[:k]
\right\|_F^2
\right)^{1/2},
\qquad
v\in\mathcal{A}_k.
\label{eq:anchor_distance}
\end{equation}
The distances are converted into normalized gate weights:
\begin{equation}
g_v
=
\frac{\exp(-d_v^{\mathrm{gate}}/T)}
{\sum_{u\in\mathcal{A}_k}\exp(-d_u^{\mathrm{gate}}/T)},
\qquad T=1.
\label{eq:anchor_gate_weight}
\end{equation}
Reliability is determined from their normalized entropy:
\begin{equation}
\overline{H}(\mathbf{g})
=
\frac{
-\sum_{v\in\mathcal{A}_k}g_v\log_2 g_v
}{
\log_2|\mathcal{A}_k|
}.
\label{eq:anchor_entropy}
\end{equation}
Low entropy indicates a distinctive anchor match, whereas high entropy indicates no reliable match.
KVCMAS accepts correction when $\overline{H}(\mathbf{g})\leq\tau$; thus, a higher $\tau$ increases the realized reuse ratio $\rho$.
Fewer than two compatible anchors trigger dense prefill.
Appendix~\ref{app:exp_th} evaluates the resulting accuracy and reuse trade-off.

\paragraph{Correction Weighting and Positional Alignment.}
After accepting reuse, KVCMAS computes token-wise interpolation weights from the mean absolute source cache distance, separately for keys and values, and applies a softmax with $T=1$.
Equation~\ref{eq:kvcmas_delta} omits these token and key/value indices for readability.
For key matching, the source and anchor keys are represented in the same canonical positional coordinates before their distances are computed.
The corrected keys are then aligned to their target positions, while values require no rotation.
These operations proceed layer by layer without an additional full-cache copy.

\paragraph{Variable Segment Lengths.}
Anchors retain their original lengths without padding or length-specific pools.
The gate compares eligible anchors over the prefix of length $k$, while correction of a length-$S$ segment uses anchors covering at least $S$ tokens.
If no anchor covers the full segment, KVCMAS corrects the longest covered prefix and retains the position re-aligned source cache for the remaining rows, avoiding extrapolation beyond observed anchor coverage.

\paragraph{Cache Materialization and Pool Updates.}
An accepted correction materializes the full-dimensional target cache, which becomes the source cache for the next workflow edge.
Because interpolation quality varies across transitions, KVCMAS applies the reliability gate independently at every edge.
An unreliable match triggers dense prefill, producing an exact target cache and resetting propagated cache error.
After dense prefill, KVCMAS computes:
\begin{equation}
\Delta_{j\leftarrow i,\mathrm{seg}}
=
C_{j,\mathrm{seg}}
-
\operatorname{Align}(C_{i,\mathrm{seg}}),
\label{eq:observed_transition}
\end{equation}
where $C_{i,\mathrm{seg}}$ is the materialized source cache carried by the workflow.
It factorizes the source cache and observed correction and inserts their low-rank factors as a new anchor.
Accepted corrections are not reinserted, so pool updates use target caches obtained through dense model execution.
Following KVComm, the least frequently used anchor is evicted among the five oldest entries when the pool is full.

\newpage
\section{Main Experiments}
\label{app:exp}

\subsection{Experimental Setup}
\label{app:exp_setup}

As described in Section~\ref{sec:exp_setup}, we build our evaluation on the multi-agent framework of KVComm~\citep{ye2025kvcomm}.
Table~\ref{tab:exp_setup} summarizes the agent family, number of evaluated samples, and reliability threshold used for each workload.

\begin{table*}[h]
\centering
\caption{
Detailed workload configuration used in the accuracy evaluation.
MathVista and Video-MME inherit the agent prompt structures of GSM8K and MMLU, respectively.
}
\label{tab:exp_setup}
\resizebox{0.95\textwidth}{!}{
\begin{tabular}{lccccc}
\toprule
& \textbf{MMLU} & \textbf{GSM8K} & \textbf{HumanEval} & \textbf{MathVista} & \textbf{Video-MME} \\
\midrule
\textbf{Agent Family} & AnalyzeAgent & MathSolver & CodeWriting & MathSolver & AnalyzeAgent \\
\textbf{\# Samples} & 1,531 & 1,319 & 161 & 1,000 & 1,296 \\
\textbf{Threshold $\tau$} & 0.29 & 0.32 & 0.34 & 0.50 & 0.60 \\
\bottomrule
\end{tabular}
}
\end{table*}

\textbf{Agent Roles.}
Following KVComm~\citep{ye2025kvcomm}, we use three task-specific agents followed by a \textit{FinalRefer} reflection agent, which reflects on the preceding agent outputs and produces the final answer.
MMLU uses a \textit{Knowledgeable Expert}, \textit{Wiki Searcher}, and \textit{Critic} to identify relevant entities, reason over retrieved information, and examine preceding analyses.
GSM8K uses a \textit{Math Solver}, \textit{Mathematical Analyst}, and \textit{Programming Expert} for complementary mathematical solutions, while HumanEval uses a \textit{Project Manager}, \textit{Algorithm Designer}, and \textit{Programming Expert} for code planning and implementation.
The \textit{FinalRefer} agent produces the final answer from their outputs.
MathVista and Video-MME retain the GSM8K and MMLU role structures, respectively, with prompts adapted to the provided image or video.

\textbf{Workflow Structures.}
The correction procedure applies to sequential, loop, fan-in, and fan-out transitions.
At fan-in, each incoming segment retains its source cache and is corrected separately.
Dense fallback produces an exact target cache for the current segment and resets its propagated cache error for subsequent transitions.
Appendices~\ref{app:abl_agentn} and~\ref{app:abl_round} evaluate varying numbers of agents and interaction rounds.

\textbf{Benchmark Samples and VLM Inputs.}
Each configuration is evaluated three times, with accuracy variation reported in Appendix~\ref{app:exp_acc}.
For MathVista, LLaVA-OneVision uses its any-resolution input path, producing visual context lengths that vary with image dimensions.
For Video-MME, we sample 32 frames, resize each to $384\times384$, and pool each frame to 196 visual tokens.

\textbf{Reliability Threshold.}
We select $\tau$ to obtain approximately $\rho=0.8$ for the LLM workloads and $\rho=0.6$ for the VLM workloads.
Starting from the $\tau=0.3$ default of KVComm~\citep{ye2025kvcomm}, we use $\tau=0.29$, $0.32$, and $0.34$ for MMLU, GSM8K, and HumanEval, respectively, and $\tau=0.50$ and $0.60$ for MathVista and Video-MME.
Appendix~\ref{app:exp_th} reports the resulting accuracy and reuse trade-off.

\textbf{Anchor Pool Size.}
KVComm uses $V=20$ as its original balance between accuracy and efficiency.
KVCMAS uses $V=10$, which captures most of the accuracy benefit with a smaller pool; larger pools provide only marginal gains with additional memory overhead (Appendix~\ref{app:abl_anchor}).

\textbf{Factorization and Measurement.}
After dense fallback, KVCMAS computes a rank-$r$ factorization for each layer's keys and values.
Anchor construction occurs after the corresponding hop generates its first token and is therefore excluded from that hop's TTFT by definition.
Its execution time is included in E2E latency and throughput, and peak GPU memory includes the transient factorization state.

\textbf{Baseline Implementations.}
We implement all baselines within the same evaluation framework and apply the same key position alignment before cache reuse.
GraphFlow is implemented as the representative preconstructed correction baseline: each correction is calibrated on the corresponding recurring context relation relative to a context-free reference and applied to subsequent requests without online matching.
This preserves its correction construction and application while excluding graph-specific scheduling components outside the scope of cross-agent KV cache correction.
KVComm follows its original online anchor matching procedure with $V=20$, while the selective recomputation baselines follow the layer and token selection criteria described in Section~\ref{sec:background_methods}.

\newpage
\subsection{Accuracy Evaluation}
\label{app:exp_acc}

Section~\ref{sec:exp_acc} reports the mean accuracy and latency measured on the trajectories generated by each method.
Here, we examine the trajectory lengths covered by the evaluation and the accuracy variation across three runs.
Table~\ref{tab:trajectory_length} reports the average and maximum accumulated trajectory lengths in model tokens, including visual tokens for VLM workloads.

\begin{table*}[h]
\centering
\caption{
Average and maximum trajectory lengths during the accuracy evaluation, reported in model tokens.
Visual tokens are included for MathVista and Video-MME.
}
\label{tab:trajectory_length}
\resizebox{\textwidth}{!}{
\begin{tabular}{lcccccccccc}
\toprule
& \multicolumn{2}{c}{MMLU}
& \multicolumn{2}{c}{GSM8K}
& \multicolumn{2}{c}{HumanEval}
& \multicolumn{2}{c}{MathVista}
& \multicolumn{2}{c}{Video-MME} \\
\cmidrule(lr){2-3}
\cmidrule(lr){4-5}
\cmidrule(lr){6-7}
\cmidrule(lr){8-9}
\cmidrule(lr){10-11}
Method
& Average & Maximum
& Average & Maximum
& Average & Maximum
& Average & Maximum
& Average & Maximum \\
\midrule
NonShared    & 2,324 & 3,176 & 2,267 & 3,460 & 1,809 & 3,277 & 4,216 & 9,032 & 6,594 & 7,659 \\
FullShared   & 1,986 & 2,947 & 2,597 & 3,911 & 1,774 & 3,563 & 4,199 & 8,904 & 7,666 & 9,194 \\
DroidSpeak   & 2,115 & 3,116 & 2,312 & 3,965 & 1,738 & 3,434 & 4,223 & 9,033 & 7,415 & 9,150 \\
CacheBlend   & 2,345 & 3,432 & 2,275 & 3,919 & 1,774 & 3,104 & 4,213 & 9,033 & 6,943 & 9,206 \\
RelayCaching & 2,169 & 3,851 & 2,105 & 3,994 & 1,672 & 3,155 & 4,282 & 9,414 & 8,745 & 9,276 \\
GraphFlow    & 2,461 & 3,201 & 2,670 & 4,438 & 1,866 & 3,519 & 4,230 & 8,904 & 6,645 & 7,657 \\
KVComm       & 2,350 & 3,362 & 2,394 & 3,975 & 1,802 & 3,283 & 4,215 & 8,971 & 6,601 & 7,763 \\
KVCMAS       & 2,253 & 3,362 & 2,312 & 3,525 & 1,818 & 3,404 & 4,219 & 9,036 & 6,733 & 8,078 \\
\bottomrule
\end{tabular}
}
\end{table*}

The maximum lengths substantially exceed the averages for several workloads, particularly MathVista, showing that the evaluation includes inputs with long accumulated context.
These samples cover the long-context regime in which the serving advantage of KVCMAS increases, as shown in Section~\ref{sec:exp_eff}.

Table~\ref{tab:accuracy_deviation} reports the standard deviation of accuracy in percentage points.
KVCMAS shows standard deviations between $0.3\%$ and $0.9\%$ points, comparable to the other correction methods.

\begin{table*}[h]
\centering
\caption{
Standard deviation of accuracy across three runs, reported in percentage points.
}
\label{tab:accuracy_deviation}
\begin{tabular}{lccccc}
\toprule
Method & MMLU & GSM8K & HumanEval & MathVista & Video-MME \\
\midrule
NonShared    & 0.26 & 0.22 & 0.59 & 0.35 & 0.37 \\
FullShared   & 0.27 & 0.28 & 0.66 & 0.35 & 0.36 \\
DroidSpeak   & 0.43 & 0.46 & 1.03 & 0.54 & 0.58 \\
CacheBlend   & 0.43 & 0.40 & 1.03 & 0.55 & 0.58 \\
RelayCaching & 0.43 & 0.40 & 1.01 & 0.55 & 0.57 \\
GraphFlow    & 0.37 & 0.38 & 1.14 & 0.46 & 0.45 \\
KVComm       & 0.37 & 0.33 & 0.80 & 0.47 & 0.50 \\
KVCMAS       & 0.36 & 0.35 & 0.86 & 0.49 & 0.47 \\
\bottomrule
\end{tabular}
\end{table*}

\newpage
\subsection{Threshold}
\label{app:exp_th}

Section~\ref{sec:exp_setup} selects the reliability threshold $\tau$ to obtain the target reuse regime for each workload.
Here, we vary $\tau$ on representative LLM and VLM workloads, MMLU and Video-MME, and measure accuracy and the realized reuse ratio $\rho$.
As defined in Appendix~\ref{app:obs_anchor}, $\tau$ is applied to the normalized entropy of the anchor weights.
A higher $\tau$ accepts a broader set of corrections and generally increases $\rho$.

\begin{table}[h]
\centering
\caption{
Effect of $\tau$ on accuracy (\%) and reuse ratio.
The settings used in the main evaluation are shown in \textbf{bold}.
}
\label{tab:threshold}
\begin{tabular}{ccccccc}
\toprule
\multicolumn{3}{c}{MMLU}
&
\multicolumn{3}{c}{Video-MME} \\
\cmidrule(lr){1-3}
\cmidrule(lr){4-6}
$\tau$ & Accuracy & $\rho$
& $\tau$ & Accuracy & $\rho$ \\
\midrule
0.10 & 67.47 & 0.623 & 0.20 & 52.98 & 0.590 \\
0.20 & 66.82 & 0.705 & 0.40 & 52.88 & 0.610 \\
0.25 & 66.62 & 0.720 & 0.50 & 52.80 & 0.630 \\
0.27 & 66.82 & 0.762 & 0.55 & 52.64 & 0.630 \\
\textbf{0.29} & \textbf{65.45} & \textbf{0.790} & 0.58 & 52.44 & 0.630 \\
0.30 & 64.66 & 0.820 & \textbf{0.60} & \textbf{52.36} & \textbf{0.641} \\
0.31 & 64.47 & 0.875 & 0.62 & 52.14 & 0.641 \\
0.33 & 64.01 & 0.886 & 0.65 & 52.04 & 0.650 \\
0.35 & 61.92 & 0.905 & 0.70 & 51.96 & 0.650 \\
0.40 & 59.96 & 0.939 & 0.80 & 51.95 & 0.649 \\
\bottomrule
\end{tabular}
\end{table}

Table~\ref{tab:threshold} shows that $\tau$ controls the accuracy and reuse trade-off.
Increasing the threshold generally raises $\rho$ while gradually lowering accuracy.
The selected thresholds reach the target reuse regimes without entering the more aggressive range where additional reuse causes larger accuracy loss.

On Video-MME, one of the three reuse-eligible transitions exhibits consistently large KV cache deviation and is routed to dense prefill throughout the threshold sweep.
Consequently, $\rho$ saturates near 0.65, close to the attainable reuse level when the other two transitions reuse their caches.
Unlike fixed-budget selective recomputation, KVCMAS evaluates each transition independently and avoids forcing reuse when anchor correction is unreliable.

\newpage
\subsection{Serving Efficiency Evaluations}
\label{app:exp_serve}

\begin{figure*}[h]
    \centering
    \includegraphics[width=\textwidth]{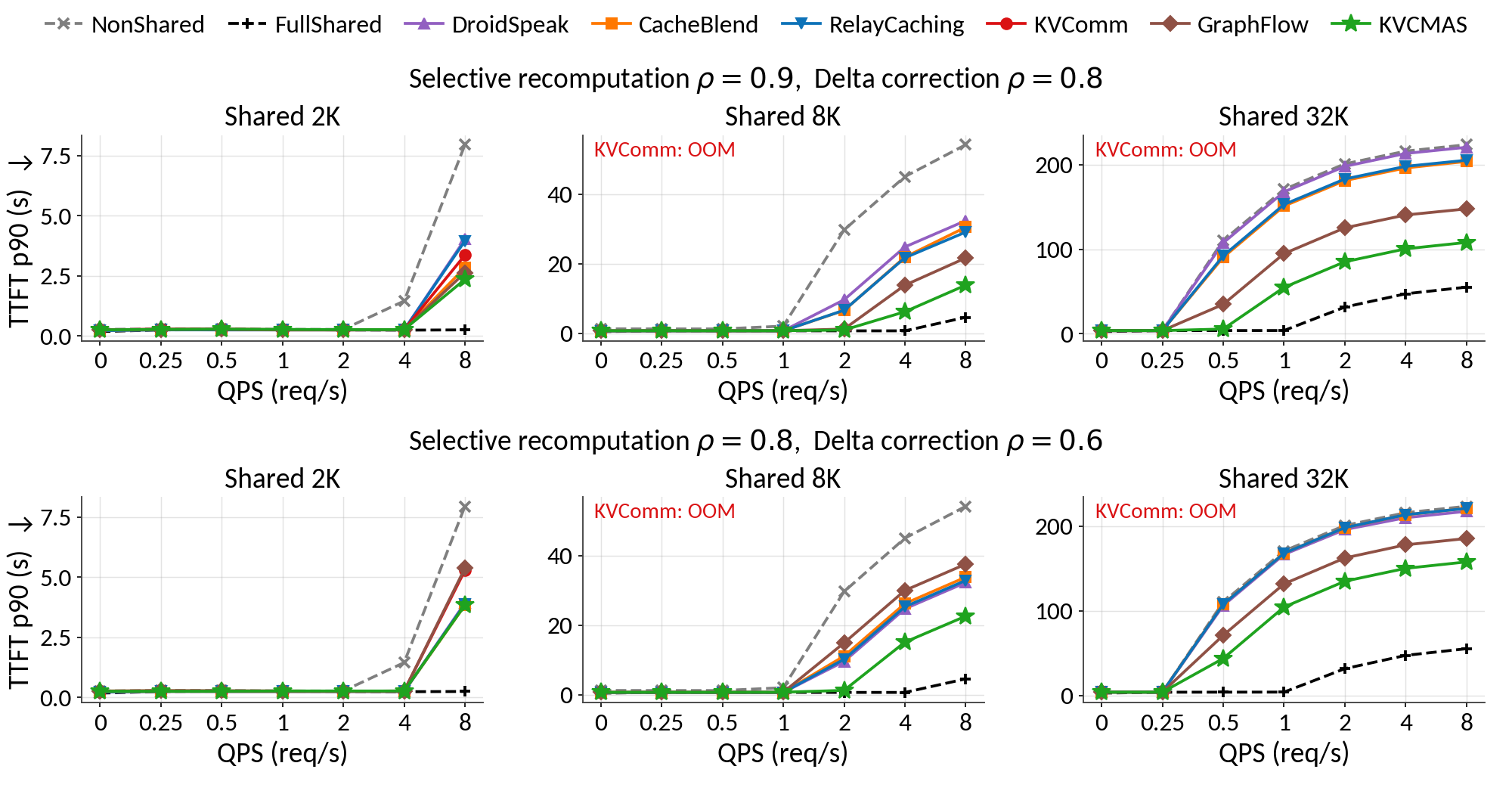}
    \caption{
    Tail (p90) TTFT under concurrent serving across shared context lengths and QPS.
    The reuse ratio configurations follow Figure~\ref{fig:serve_eff}.
    }
    \label{fig:serve_eff_p90_app}
\end{figure*}

Figure~\ref{fig:serve_eff} in Section~\ref{sec:exp_eff} summarizes median TTFT under concurrent serving.
Here, we report the complete p50 and p90 values across shared context lengths and request rates, together with the p90 trends in Figure~\ref{fig:serve_eff_p90_app}.
The experiments use Llama-3.1-8B on an NVIDIA A100 80\,GB GPU.
We replay agent inputs with accumulated 128-token outputs and execute each request through its first generated token to measure TTFT.
We compare selective recomputation at $\rho=0.9$ and $0.8$ with delta correction at $\rho=0.8$ and $0.6$, respectively. QPS 0 denotes single-stream request execution without overlapping arrivals.

Table~\ref{tab:app_serve} reports the corresponding numerical results.
At 32K shared tokens and 8 QPS under the $\rho=0.8$ delta correction setting, KVCMAS provides 2.0$\times$ and 2.1$\times$ p50 and p90 TTFT speedups over NonShared, respectively.
It also reduces both p50 and p90 TTFT by 27\% relative to GraphFlow, the next-fastest KV cache sharing work.
The advantage increases with shared context length, showing the efficiency of KVCMAS for long-context workloads under high request rates.
KVComm exceeds GPU memory capacity at 8K and 32K because it stores full-dimensional anchor states.

\newpage
\begin{table*}[h]
\centering
\caption{
Median (p50) and tail (p90) TTFT in seconds under concurrent serving across shared context lengths, request rates, and reuse ratios $\rho$.
}
\label{tab:app_serve}
\setlength{\tabcolsep}{1.8pt}
\scriptsize
\textbf{(a) Shared context of 2K}\\[2pt]
\resizebox{\textwidth}{!}{
\begin{tabular}{clrrrrrrrrrrrrrrrrrrrrr}
\toprule
& &
\multicolumn{7}{c}{$\rho=0.9$} &
\multicolumn{7}{c}{$\rho=0.8$} &
\multicolumn{7}{c}{$\rho=0.6$} \\
\cmidrule(lr){3-9}\cmidrule(lr){10-16}\cmidrule(lr){17-23}
\textbf{Pct.} & \textbf{Method}
& 0 & .25 & .5 & 1 & 2 & 4 & 8
& 0 & .25 & .5 & 1 & 2 & 4 & 8
& 0 & .25 & .5 & 1 & 2 & 4 & 8 \\
\midrule
p50 & NonShared    & 0.26 & 0.27 & 0.26 & 0.26 & 0.26 & 0.57 & 4.66 & 0.26 & 0.27 & 0.26 & 0.26 & 0.26 & 0.57 & 4.66 & 0.26 & 0.27 & 0.26 & 0.26 & 0.26 & 0.57 & 4.66 \\
    & FullShared   & 0.03 & 0.03 & 0.03 & 0.03 & 0.03 & 0.03 & 0.10 & 0.03 & 0.03 & 0.03 & 0.03 & 0.03 & 0.03 & 0.10 & 0.03 & 0.03 & 0.03 & 0.03 & 0.03 & 0.03 & 0.10 \\
    & DroidSpeak   & 0.17 & 0.17 & 0.17 & 0.16 & 0.16 & 0.16 & 2.84 & 0.16 & 0.17 & 0.17 & 0.16 & 0.16 & 0.16 & 2.93 & 0.16 & 0.18 & 0.17 & 0.16 & 0.16 & 0.16 & 3.99 \\
    & CacheBlend   & 0.15 & 0.16 & 0.15 & 0.14 & 0.14 & 0.14 & 2.07 & 0.17 & 0.18 & 0.17 & 0.17 & 0.17 & 0.16 & 2.87 & 0.17 & 0.18 & 0.18 & 0.17 & 0.17 & 0.17 & 3.11 \\
    & RelayCaching & 0.15 & 0.16 & 0.16 & 0.15 & 0.15 & 0.15 & 3.18 & 0.17 & 0.18 & 0.17 & 0.16 & 0.16 & 0.16 & 2.99 & 0.17 & 0.18 & 0.18 & 0.17 & 0.17 & 0.17 & 3.30 \\
    & KVComm       & 0.14 & 0.16 & 0.16 & 0.16 & 0.16 & 0.16 & 1.49 & 0.14 & 0.18 & 0.17 & 0.17 & 0.16 & 0.17 & 2.96 & 0.25 & 0.25 & 0.25 & 0.25 & 0.25 & 0.25 & 3.69 \\
    & GraphFlow    & 0.14 & 0.16 & 0.16 & 0.16 & 0.16 & 0.16 & 1.12 & 0.14 & 0.18 & 0.17 & 0.17 & 0.16 & 0.18 & 2.24 & 0.25 & 0.25 & 0.25 & 0.25 & 0.25 & 0.25 & 3.87 \\
    & \textbf{KVCMAS} & 0.14 & 0.14 & 0.14 & 0.14 & 0.14 & 0.14 & 0.28 & 0.14 & 0.15 & 0.15 & 0.15 & 0.16 & 0.16 & 2.06 & 0.25 & 0.25 & 0.25 & 0.25 & 0.25 & 0.25 & 2.81 \\
\midrule
p90 & NonShared    & 0.27 & 0.29 & 0.29 & 0.27 & 0.27 & 1.45 & 7.95 & 0.27 & 0.29 & 0.29 & 0.27 & 0.27 & 1.45 & 7.95 & 0.27 & 0.29 & 0.29 & 0.27 & 0.27 & 1.45 & 7.95 \\
    & FullShared   & 0.18 & 0.25 & 0.26 & 0.26 & 0.25 & 0.24 & 0.25 & 0.18 & 0.25 & 0.26 & 0.26 & 0.25 & 0.24 & 0.25 & 0.18 & 0.25 & 0.26 & 0.26 & 0.25 & 0.24 & 0.25 \\
    & DroidSpeak   & 0.23 & 0.26 & 0.25 & 0.25 & 0.25 & 0.25 & 4.02 & 0.22 & 0.27 & 0.25 & 0.25 & 0.25 & 0.25 & 3.93 & 0.22 & 0.27 & 0.25 & 0.25 & 0.25 & 0.25 & 5.23 \\
    & CacheBlend   & 0.22 & 0.26 & 0.26 & 0.25 & 0.25 & 0.25 & 2.84 & 0.23 & 0.25 & 0.25 & 0.25 & 0.25 & 0.25 & 3.84 & 0.23 & 0.25 & 0.25 & 0.25 & 0.25 & 0.25 & 4.17 \\
    & RelayCaching & 0.22 & 0.25 & 0.25 & 0.25 & 0.25 & 0.25 & 3.95 & 0.23 & 0.25 & 0.25 & 0.25 & 0.25 & 0.25 & 3.89 & 0.23 & 0.25 & 0.25 & 0.25 & 0.25 & 0.25 & 4.38 \\
    & KVComm       & 0.25 & 0.28 & 0.26 & 0.26 & 0.25 & 0.25 & 1.91 & 0.25 & 0.28 & 0.28 & 0.26 & 0.25 & 0.26 & 3.36 & 0.27 & 0.28 & 0.27 & 0.27 & 0.26 & 0.27 & 5.29 \\
    & GraphFlow    & 0.25 & 0.27 & 0.27 & 0.26 & 0.25 & 0.25 & 1.82 & 0.25 & 0.27 & 0.28 & 0.25 & 0.25 & 0.26 & 2.63 & 0.26 & 0.28 & 0.28 & 0.27 & 0.26 & 0.27 & 5.38 \\
    & \textbf{KVCMAS} & 0.26 & 0.28 & 0.27 & 0.26 & 0.26 & 0.25 & 1.68 & 0.25 & 0.27 & 0.28 & 0.27 & 0.26 & 0.26 & 2.35 & 0.26 & 0.26 & 0.26 & 0.27 & 0.27 & 0.27 & 3.84 \\
\bottomrule
\end{tabular}
}

\vspace{5pt}

\textbf{(b) Shared context of 8K}\\[2pt]
\resizebox{\textwidth}{!}{
\begin{tabular}{clrrrrrrrrrrrrrrrrrrrrr}
\toprule
& &
\multicolumn{7}{c}{$\rho=0.9$} &
\multicolumn{7}{c}{$\rho=0.8$} &
\multicolumn{7}{c}{$\rho=0.6$} \\
\cmidrule(lr){3-9}\cmidrule(lr){10-16}\cmidrule(lr){17-23}
\textbf{Pct.} & \textbf{Method}
& 0 & .25 & .5 & 1 & 2 & 4 & 8
& 0 & .25 & .5 & 1 & 2 & 4 & 8
& 0 & .25 & .5 & 1 & 2 & 4 & 8 \\
\midrule
p50 & NonShared    & 1.15 & 1.18 & 1.25 & 1.51 & 16.81 & 25.24 & 29.52 & 1.15 & 1.18 & 1.25 & 1.51 & 16.81 & 25.24 & 29.52 & 1.15 & 1.18 & 1.25 & 1.51 & 16.81 & 25.24 & 29.52 \\
    & FullShared   & 0.03 & 0.04 & 0.04 & 0.04 & 0.14 & 0.39 & 2.78 & 0.03 & 0.04 & 0.04 & 0.04 & 0.14 & 0.39 & 2.78 & 0.03 & 0.04 & 0.04 & 0.04 & 0.14 & 0.39 & 2.78 \\
    & DroidSpeak   & 0.64 & 0.65 & 0.64 & 0.64 & 5.89 & 14.31 & 18.51 & 0.65 & 0.65 & 0.64 & 0.64 & 5.87 & 14.28 & 18.46 & 0.65 & 0.65 & 0.65 & 0.65 & 5.99 & 14.38 & 18.62 \\
    & CacheBlend   & 0.56 & 0.57 & 0.56 & 0.56 & 4.16 & 12.66 & 17.04 & 0.65 & 0.66 & 0.66 & 0.66 & 6.36 & 14.75 & 18.97 & 0.67 & 0.67 & 0.66 & 0.67 & 6.48 & 14.81 & 19.00 \\
    & RelayCaching & 0.57 & 0.58 & 0.56 & 0.57 & 3.98 & 12.54 & 16.62 & 0.65 & 0.66 & 0.65 & 0.65 & 5.87 & 14.25 & 18.45 & 0.66 & 0.67 & 0.66 & 0.67 & 6.55 & 14.85 & 19.07 \\
    & GraphFlow    & 0.38 & 0.65 & 0.64 & 0.65 & 4.55 & 5.96 & 10.09 & 0.38 & 0.67 & 0.66 & 0.66 & 0.98 & 8.49 & 12.56 & 0.75 & 0.75 & 0.74 & 0.75 & 8.59 & 17.35 & 21.55 \\
    & \textbf{KVCMAS} & 0.25 & 0.26 & 0.26 & 0.28 & 0.28 & 1.47 & 5.65 & 0.38 & 0.50 & 0.50 & 0.52 & 0.55 & 3.94 & 8.15 & 0.72 & 0.73 & 0.73 & 0.73 & 0.98 & 8.96 & 13.04 \\
\midrule
p90 & NonShared    & 1.29 & 1.27 & 1.27 & 2.08 & 29.70 & 45.03 & 54.19 & 1.29 & 1.27 & 1.27 & 2.08 & 29.70 & 45.03 & 54.19 & 1.29 & 1.27 & 1.27 & 2.08 & 29.70 & 45.03 & 54.19 \\
    & FullShared   & 0.51 & 0.72 & 0.72 & 0.73 & 0.72 & 0.72 & 4.61 & 0.51 & 0.72 & 0.72 & 0.73 & 0.72 & 0.72 & 4.61 & 0.51 & 0.72 & 0.72 & 0.73 & 0.72 & 0.72 & 4.61 \\
    & DroidSpeak   & 0.70 & 0.74 & 0.72 & 0.72 & 9.64 & 24.82 & 32.37 & 0.70 & 0.74 & 0.72 & 0.72 & 9.63 & 24.75 & 32.29 & 0.70 & 0.72 & 0.73 & 0.73 & 9.84 & 24.95 & 32.57 \\
    & CacheBlend   & 0.68 & 0.75 & 0.72 & 0.72 & 6.59 & 21.89 & 30.49 & 0.71 & 0.75 & 0.73 & 0.74 & 11.15 & 26.24 & 33.83 & 0.71 & 0.74 & 0.72 & 0.72 & 10.81 & 25.84 & 33.40 \\
    & RelayCaching & 0.68 & 0.73 & 0.72 & 0.72 & 6.64 & 21.71 & 29.09 & 0.70 & 0.73 & 0.72 & 0.72 & 10.24 & 25.34 & 32.89 & 0.70 & 0.73 & 0.72 & 0.72 & 10.91 & 25.91 & 33.50 \\
    & GraphFlow    & 0.74 & 0.75 & 0.74 & 0.75 & 6.55 & 9.49 & 17.09 & 0.74 & 0.77 & 0.75 & 0.76 & 1.23 & 13.84 & 21.65 & 0.76 & 0.78 & 0.76 & 0.76 & 14.95 & 30.02 & 37.58 \\
    & \textbf{KVCMAS} & 0.74 & 0.74 & 0.74 & 0.74 & 0.75 & 1.93 & 9.15 & 0.74 & 0.75 & 0.74 & 0.75 & 0.96 & 6.21 & 13.84 & 0.76 & 0.76 & 0.76 & 0.78 & 1.31 & 15.16 & 22.63 \\
\bottomrule
\end{tabular}
}
\vspace{5pt}

\textbf{(c) Shared context of 32K}\\[2pt]
\resizebox{\textwidth}{!}{
\begin{tabular}{clrrrrrrrrrrrrrrrrrrrrr}
\toprule
& &
\multicolumn{7}{c}{$\rho=0.9$} &
\multicolumn{7}{c}{$\rho=0.8$} &
\multicolumn{7}{c}{$\rho=0.6$} \\
\cmidrule(lr){3-9}\cmidrule(lr){10-16}\cmidrule(lr){17-23}
\textbf{Pct.} & \textbf{Method}
& 0 & .25 & .5 & 1 & 2 & 4 & 8
& 0 & .25 & .5 & 1 & 2 & 4 & 8
& 0 & .25 & .5 & 1 & 2 & 4 & 8 \\
\midrule
p50 & NonShared    & 3.76 & 3.77 & 62.98 & 96.55 & 113.28 & 121.66 & 125.89 & 3.76 & 3.77 & 62.98 & 96.55 & 113.28 & 121.66 & 125.89 & 3.76 & 3.77 & 62.98 & 96.55 & 113.28 & 121.66 & 125.89 \\
    & FullShared   & 0.07 & 0.08 & 0.94 & 2.41 & 18.69 & 27.07 & 31.25 & 0.07 & 0.08 & 0.94 & 2.41 & 18.69 & 27.07 & 31.25 & 0.07 & 0.08 & 0.94 & 2.41 & 18.69 & 27.07 & 31.25 \\
    & DroidSpeak   & 3.66 & 3.69 & 61.32 & 94.92 & 111.71 & 120.13 & 124.17 & 3.65 & 3.66 & 60.63 & 94.23 & 110.86 & 118.35 & 122.50 & 3.65 & 3.66 & 60.12 & 93.64 & 110.58 & 119.00 & 123.28 \\
    & CacheBlend   & 3.30 & 3.31 & 51.59 & 85.37 & 102.11 & 110.43 & 114.72 & 3.68 & 3.70 & 61.73 & 95.00 & 111.81 & 120.28 & 124.64 & 3.73 & 3.73 & 61.37 & 94.94 & 111.89 & 120.29 & 124.69 \\
    & RelayCaching & 3.35 & 3.35 & 52.60 & 86.33 & 103.17 & 111.51 & 115.40 & 3.67 & 3.69 & 61.50 & 94.99 & 111.83 & 120.32 & 124.55 & 3.72 & 3.73 & 61.42 & 95.15 & 112.57 & 120.09 & 124.61 \\
    & GraphFlow    & 0.91 & 3.70 & 9.62 & 43.77 & 60.95 & 69.78 & 74.10 & 1.90 & 3.70 & 21.96 & 55.62 & 72.40 & 80.76 & 84.84 & 3.73 & 3.75 & 39.95 & 73.52 & 90.45 & 99.15 & 103.14 \\
    & \textbf{KVCMAS} & 0.91 & 0.92 & 2.27 & 18.05 & 36.02 & 44.44 & 48.62 & 1.02 & 1.90 & 3.78 & 33.52 & 49.75 & 57.93 & 61.95 & 3.75 & 3.74 & 26.70 & 60.29 & 77.20 & 85.63 & 89.84 \\
\midrule
p90 & NonShared    & 3.79 & 3.80 & 110.37 & 170.76 & 200.93 & 216.00 & 223.59 & 3.79 & 3.80 & 110.37 & 170.76 & 200.93 & 216.00 & 223.59 & 3.79 & 3.80 & 110.37 & 170.76 & 200.93 & 216.00 & 223.59 \\
    & FullShared   & 2.64 & 3.75 & 3.81 & 3.89 & 31.59 & 47.30 & 55.20 & 2.64 & 3.75 & 3.81 & 3.89 & 31.59 & 47.30 & 55.20 & 2.64 & 3.75 & 3.81 & 3.89 & 31.59 & 47.30 & 55.20 \\
    & DroidSpeak   & 3.73 & 3.81 & 107.55 & 167.86 & 197.98 & 213.25 & 220.54 & 3.73 & 3.77 & 106.22 & 166.66 & 196.07 & 210.09 & 217.64 & 3.72 & 3.77 & 105.23 & 165.67 & 196.04 & 211.18 & 219.40 \\
    & CacheBlend   & 3.63 & 3.80 & 90.42 & 151.05 & 181.16 & 196.14 & 203.79 & 3.75 & 3.81 & 108.09 & 168.09 & 198.31 & 213.46 & 221.31 & 3.75 & 3.75 & 107.51 & 168.02 & 198.48 & 213.60 & 221.80 \\
    & RelayCaching & 3.64 & 3.79 & 92.19 & 152.91 & 183.05 & 198.00 & 205.23 & 3.74 & 3.80 & 107.76 & 168.04 & 198.35 & 213.63 & 221.17 & 3.74 & 3.75 & 107.89 & 168.77 & 199.22 & 213.26 & 221.72 \\
    & GraphFlow    & 3.78 & 3.79 & 13.57 & 72.88 & 103.17 & 118.78 & 126.39 & 3.75 & 3.78 & 34.80 & 94.83 & 125.30 & 140.51 & 147.71 & 3.78 & 3.81 & 71.03 & 131.88 & 162.49 & 178.02 & 185.51 \\
    & \textbf{KVCMAS} & 3.77 & 3.77 & 4.10 & 29.01 & 59.53 & 74.82 & 82.44 & 3.77 & 3.78 & 5.64 & 54.89 & 85.18 & 100.48 & 108.05 & 3.79 & 3.80 & 43.67 & 104.45 & 134.98 & 150.22 & 157.82 \\
\bottomrule
\end{tabular}
}
\end{table*}

\newpage
\subsection{Chained Correction Efficiency Evaluations}
\label{app:exp_chaineff}

Figure~\ref{fig:chain_eff} in Section~\ref{sec:exp_eff} compares chained correction with a non-chained variant using a separate context-free reference while keeping the remaining KVCMAS configuration fixed.
Here, we report accuracy and numerical p50 and p90 TTFT across shared context lengths and request rates.
Both variants use rank $r=32$ and $V=10$ anchors.
Serving efficiency is evaluated at $\rho=0.8$.

Table~\ref{tab:chain_acc_app} shows that chained correction improves accuracy on both evaluated workloads, by $6.44\%$ points on MMLU and 1.53 points on Video-MME.
Chaining preserves an exact first-agent cache and uses each materialized cache for the next transition, preventing first-agent correction error from entering downstream shared context.

\begin{table}[h]
\vspace{-10pt}
\centering
\caption{
Accuracy of chained and non-chained KVCMAS.
The correction representation, anchor pool, reliability threshold, and reuse ratio follow the main accuracy evaluation settings.
}
\label{tab:chain_acc_app}
\begin{tabular}{lcc}
\toprule
\textbf{Reference structure} & \textbf{MMLU} & \textbf{Video-MME} \\
\midrule
\textbf{Chained} & \textbf{65.45} & \textbf{52.36} \\
Non-chained & 59.01 & 50.83 \\
\bottomrule
\end{tabular}
\end{table}

\begin{figure*}[h]
    \centering
    \includegraphics[width=\textwidth]{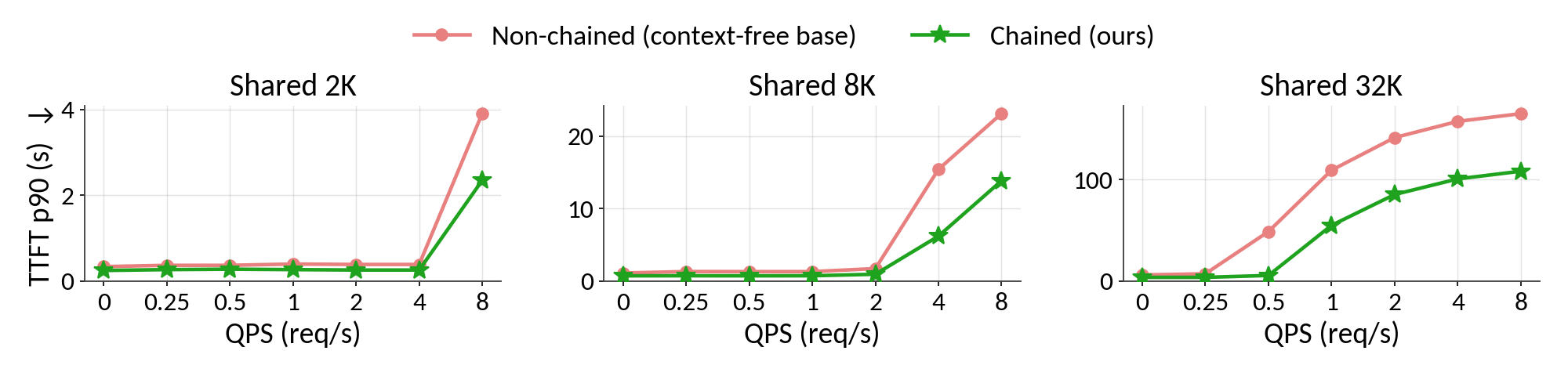}
    \vspace{-20pt}
    \caption{
    Tail (p90) TTFT of chained and non-chained correction across shared context lengths and request rates at $\rho=0.8$.
    Chained correction reduces tail latency by avoiding the separate context-free reference prefill.
    }
    \label{fig:chain_eff_p90_app}
    \vspace{-10pt}
\end{figure*}

Figure~\ref{fig:chain_eff_p90_app} reports p90 TTFT, and Table~\ref{tab:chain_eff_app} provides the numerical p50 and p90 results.
By removing the separate reference prefill, chained correction reduces both median and tail TTFT.
At 32K shared tokens and 8 QPS, it reduces both by 34\%, with larger benefits at longer shared contexts and higher request rates.

\begin{table*}[h]
\centering
\caption{
Median (p50) and tail (p90) TTFT of chained and non-chained correction across shared context lengths and request rates at $\rho=0.8$.
All values are in seconds.
}
\label{tab:chain_eff_app}
\setlength{\tabcolsep}{4.0pt}
\begin{tabular}{cllrrrrrrr}
\toprule
\multirow{2}{*}{\textbf{Pct.}}
& \multirow{2}{*}{\textbf{Shared}}
& \multirow{2}{*}{\textbf{Method}}
& \multicolumn{7}{c}{\textbf{QPS}} \\
\cmidrule(lr){4-10}
& & & \textbf{0} & \textbf{0.25} & \textbf{0.5}
& \textbf{1} & \textbf{2} & \textbf{4} & \textbf{8} \\
\midrule

\multirow{6}{*}{p50}
& \multirow{2}{*}{2K}
& Chained     & 0.14 & 0.15 & 0.15 & 0.15 & 0.16 & 0.16 & 2.06 \\
& & Non-chained & 0.15 & 0.15 & 0.15 & 0.15 & 0.15 & 0.16 & 3.42 \\
\cmidrule(lr){2-10}
& \multirow{2}{*}{8K}
& Chained     & 0.38 & 0.50 & 0.50 & 0.52 & 0.55 & 3.94 & 8.15 \\
& & Non-chained & 0.49 & 0.55 & 0.55 & 0.58 & 1.14 & 9.16 & 13.32 \\
\cmidrule(lr){2-10}
& \multirow{2}{*}{32K}
& Chained     & 1.02 & 1.90 & 3.78 & 33.52 & 49.75 & 57.93 & 61.95 \\
& & Non-chained & 1.92 & 3.78 & 29.99 & 64.07 & 81.07 & 89.77 & 93.86 \\

\midrule

\multirow{6}{*}{p90}
& \multirow{2}{*}{2K}
& Chained     & 0.25 & 0.27 & 0.28 & 0.27 & 0.26 & 0.26 & 2.35 \\
& & Non-chained & 0.34 & 0.37 & 0.37 & 0.40 & 0.39 & 0.39 & 3.91 \\
\cmidrule(lr){2-10}
& \multirow{2}{*}{8K}
& Chained     & 0.74 & 0.75 & 0.74 & 0.75 & 0.96 & 6.21 & 13.84 \\
& & Non-chained & 1.13 & 1.33 & 1.32 & 1.32 & 1.74 & 15.41 & 23.09 \\
\cmidrule(lr){2-10}
& \multirow{2}{*}{32K}
& Chained     & 3.77 & 3.78 & 5.64 & 54.89 & 85.18 & 100.48 & 108.05 \\
& & Non-chained & 6.17 & 7.28 & 48.76 & 109.05 & 141.09 & 156.99 & 164.61 \\

\bottomrule
\end{tabular}
\end{table*}

\newpage
\subsection{Serving Efficiency with Long Agent Outputs}
\label{app:exp_dec}

\begin{figure*}[h]
    \centering
    \includegraphics[width=\textwidth]{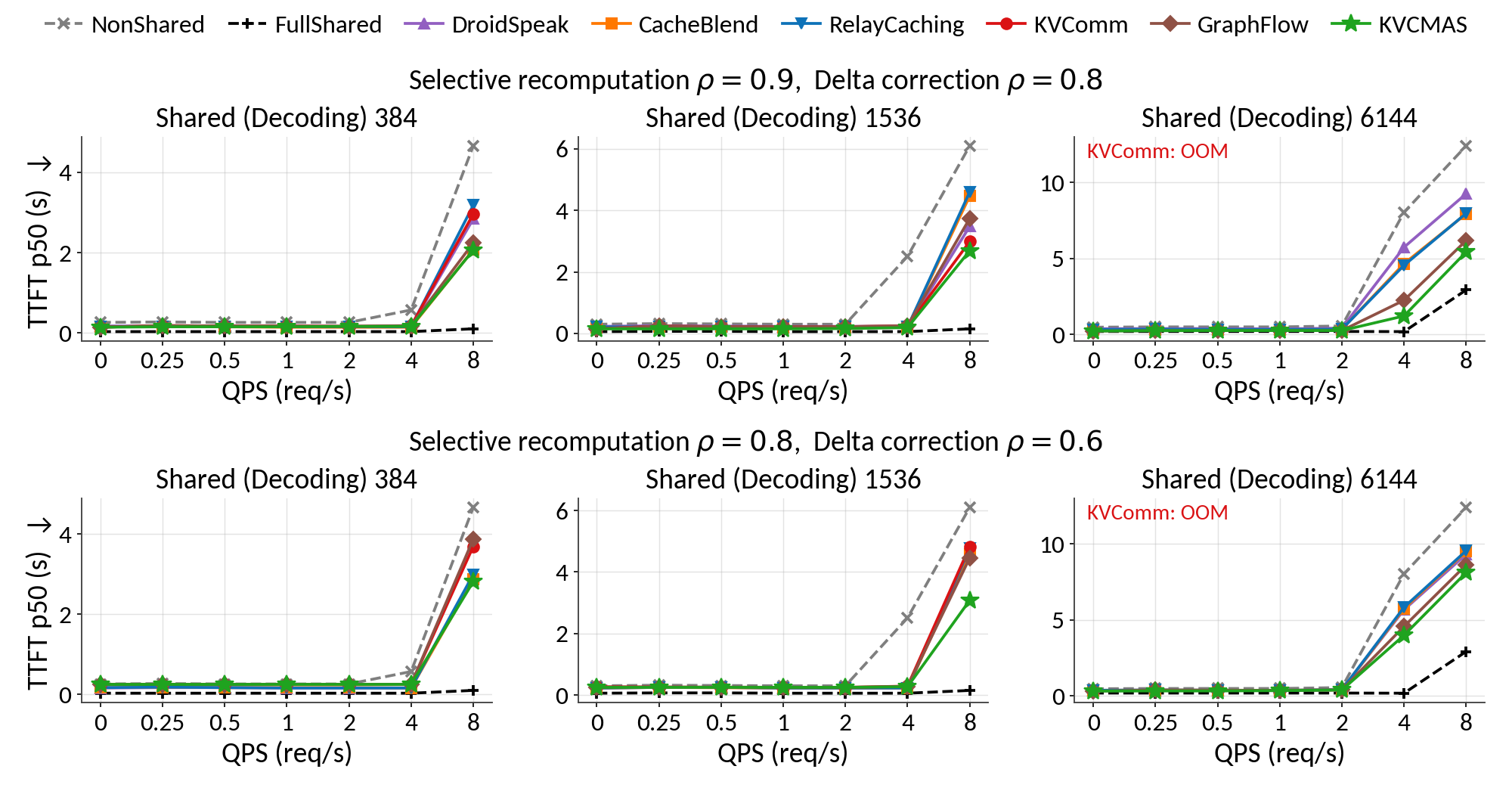}
    \vspace{-25pt}
    \caption{
    Median (p50) TTFT across request rates with 0.375K, 1.5K, and 6K shared generated tokens.
    The non-generated shared context is fixed to 2K tokens.
    Selective recomputation at $\rho=0.9$ and $0.8$ is matched with delta correction at $\rho=0.8$ and $0.6$, respectively.
    }
    \label{fig:serve_eff_dec_p50_app}
\end{figure*}

\begin{figure*}[h]
    \centering
    \includegraphics[width=\textwidth]{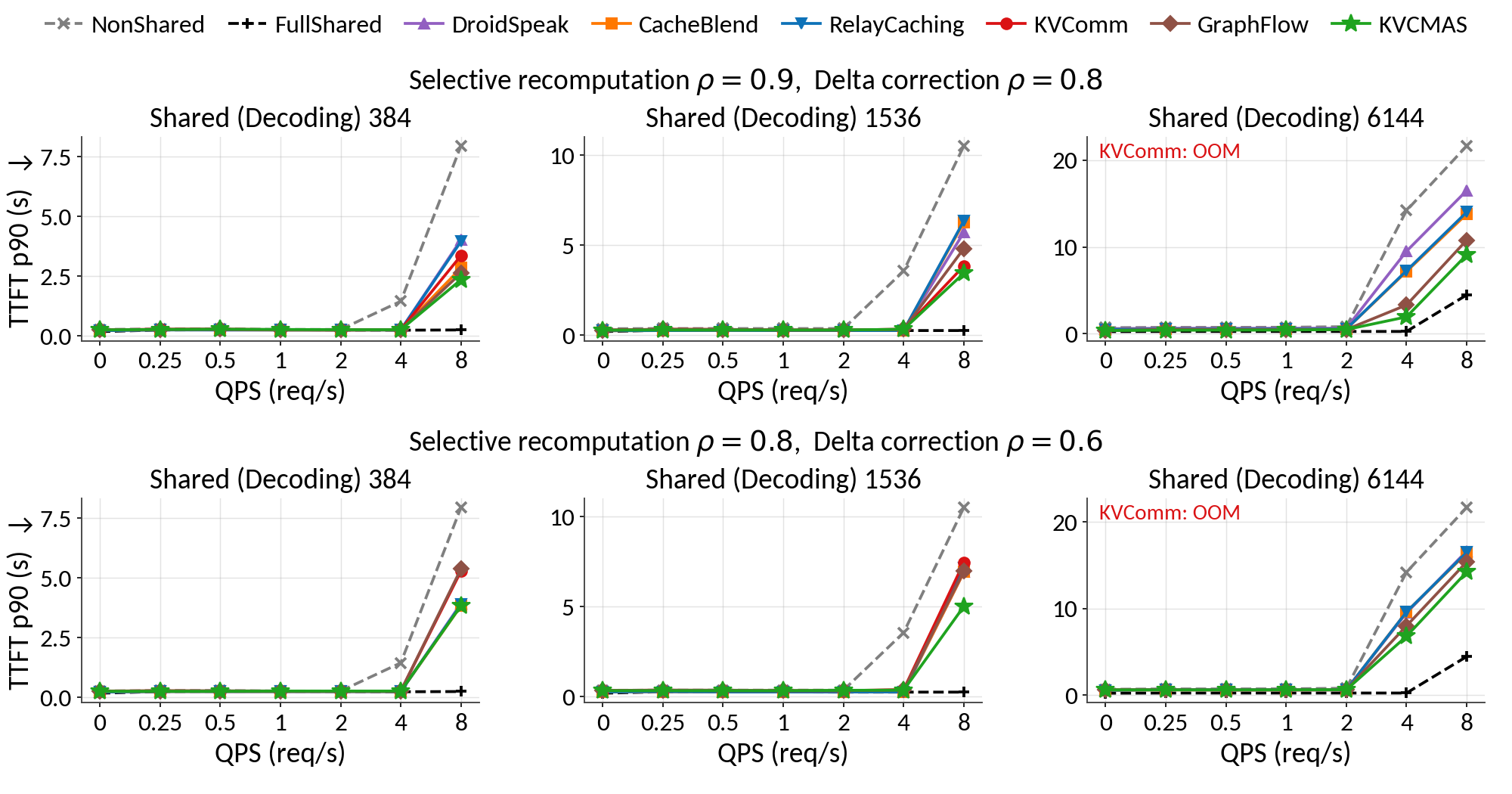}
    \vspace{-25pt}
    \caption{
    Tail (p90) TTFT under the same accumulated-output configurations as Figure~\ref{fig:serve_eff_dec_p50_app}.
    }
    \label{fig:serve_eff_dec_p90_app}
\end{figure*}

Section~\ref{sec:exp_eff} varies the non-generated shared context with 128-token agent outputs incorporated into subsequent inputs.
Here, we extend the TTFT evaluation to 0.375K, 1.5K, and 6K accumulated output tokens.
In the four-agent workflow, the final agent receives outputs from the three preceding agents, so the accumulated shared output is three times the per-agent generation length.
The 0.375K results appear in Table~\ref{tab:app_serve}(a).
Figures~\ref{fig:serve_eff_dec_p50_app} and~\ref{fig:serve_eff_dec_p90_app} show the p50 and p90 trends, while Table~\ref{tab:long_generation_app} reports the numerical results for 1.5K and 6K.

At 6K shared generated tokens and 8 QPS under the $\rho=0.8$ setting, KVCMAS provides 2.3$\times$ p50 and 2.4$\times$ p90 TTFT speedups over NonShared.
It shows the same trend at 0.375K and 1.5K, retaining the lowest TTFT among the KV cache sharing works under high request rates.
KVComm exceeds GPU memory capacity at 6K shared generated tokens because of its full-dimensional anchor pool.

\newpage
\begin{table*}[h]
\centering
\caption{
Median (p50) and tail (p90) TTFT in seconds with 1.5K and 6K shared generated tokens.
The non-generated shared context is fixed to 2K tokens.
}
\label{tab:long_generation_app}
\setlength{\tabcolsep}{1.8pt}
\scriptsize

\textbf{(a) Shared generated context: 1.5k tokens}\\[2pt]
\resizebox{\textwidth}{!}{
\begin{tabular}{clrrrrrrrrrrrrrrrrrrrrr}
\toprule
& &
\multicolumn{7}{c}{$\rho=0.9$} &
\multicolumn{7}{c}{$\rho=0.8$} &
\multicolumn{7}{c}{$\rho=0.6$} \\
\cmidrule(lr){3-9}\cmidrule(lr){10-16}\cmidrule(lr){17-23}
\textbf{Pct.} & \textbf{Method}
& 0 & .25 & .5 & 1 & 2 & 4 & 8
& 0 & .25 & .5 & 1 & 2 & 4 & 8
& 0 & .25 & .5 & 1 & 2 & 4 & 8 \\
\midrule
p50 & NonShared
& 0.30 & 0.32 & 0.31 & 0.30 & 0.30 & 2.51 & 6.10
& 0.30 & 0.32 & 0.31 & 0.30 & 0.30 & 2.51 & 6.10
& 0.30 & 0.32 & 0.31 & 0.30 & 0.30 & 2.51 & 6.10 \\
& FullShared
& 0.06 & 0.07 & 0.07 & 0.06 & 0.06 & 0.06 & 0.15
& 0.06 & 0.07 & 0.07 & 0.06 & 0.06 & 0.06 & 0.15
& 0.06 & 0.07 & 0.07 & 0.06 & 0.06 & 0.06 & 0.15 \\
& DroidSpeak
& 0.23 & 0.25 & 0.24 & 0.23 & 0.23 & 0.23 & 3.48
& 0.23 & 0.25 & 0.25 & 0.23 & 0.23 & 0.23 & 4.67
& 0.23 & 0.25 & 0.24 & 0.23 & 0.23 & 0.23 & 4.40 \\
& CacheBlend
& 0.20 & 0.21 & 0.21 & 0.20 & 0.20 & 0.20 & 4.47
& 0.23 & 0.25 & 0.24 & 0.23 & 0.23 & 0.23 & 4.58
& 0.24 & 0.25 & 0.25 & 0.23 & 0.23 & 0.24 & 3.60 \\
& RelayCaching
& 0.21 & 0.22 & 0.22 & 0.21 & 0.21 & 0.21 & 4.60
& 0.23 & 0.25 & 0.25 & 0.23 & 0.23 & 0.23 & 4.77
& 0.24 & 0.25 & 0.25 & 0.23 & 0.23 & 0.24 & 5.12 \\
& KVComm
& 0.16 & 0.19 & 0.19 & 0.19 & 0.19 & 0.19 & 2.19
& 0.15 & 0.24 & 0.22 & 0.22 & 0.22 & 0.25 & 2.99
& 0.27 & 0.26 & 0.25 & 0.25 & 0.25 & 0.29 & 4.83 \\
& GraphFlow
& 0.15 & 0.19 & 0.19 & 0.19 & 0.19 & 0.19 & 2.15
& 0.15 & 0.24 & 0.22 & 0.22 & 0.22 & 0.25 & 3.74
& 0.26 & 0.26 & 0.26 & 0.25 & 0.25 & 0.28 & 4.46 \\
& \textbf{KVCMAS}
& 0.10 & 0.10 & 0.09 & 0.09 & 0.09 & 0.17 & 1.25
& 0.15 & 0.15 & 0.15 & 0.15 & 0.16 & 0.19 & 2.69
& 0.25 & 0.25 & 0.25 & 0.25 & 0.25 & 0.27 & 3.08 \\
\midrule
p90 & NonShared
& 0.35 & 0.36 & 0.36 & 0.36 & 0.36 & 3.56 & 10.52
& 0.35 & 0.36 & 0.36 & 0.36 & 0.36 & 3.56 & 10.52
& 0.35 & 0.36 & 0.36 & 0.36 & 0.36 & 3.56 & 10.52 \\
& FullShared
& 0.19 & 0.27 & 0.26 & 0.26 & 0.25 & 0.25 & 0.25
& 0.19 & 0.27 & 0.26 & 0.26 & 0.25 & 0.25 & 0.25
& 0.19 & 0.27 & 0.26 & 0.26 & 0.25 & 0.25 & 0.25 \\
& DroidSpeak
& 0.26 & 0.28 & 0.27 & 0.26 & 0.26 & 0.26 & 5.71
& 0.26 & 0.28 & 0.27 & 0.26 & 0.26 & 0.26 & 6.93
& 0.26 & 0.27 & 0.26 & 0.26 & 0.26 & 0.26 & 6.63 \\
& CacheBlend
& 0.24 & 0.25 & 0.25 & 0.25 & 0.25 & 0.25 & 6.27
& 0.26 & 0.28 & 0.26 & 0.26 & 0.26 & 0.26 & 6.93
& 0.26 & 0.28 & 0.27 & 0.26 & 0.26 & 0.27 & 5.96 \\
& RelayCaching
& 0.24 & 0.25 & 0.26 & 0.25 & 0.25 & 0.25 & 6.35
& 0.26 & 0.28 & 0.26 & 0.26 & 0.26 & 0.26 & 7.16
& 0.26 & 0.29 & 0.27 & 0.26 & 0.26 & 0.27 & 7.59 \\
& KVComm
& 0.27 & 0.28 & 0.28 & 0.28 & 0.26 & 0.26 & 2.63
& 0.27 & 0.32 & 0.31 & 0.28 & 0.28 & 0.32 & 3.84
& 0.34 & 0.36 & 0.33 & 0.33 & 0.33 & 0.39 & 7.46 \\
& GraphFlow
& 0.27 & 0.28 & 0.28 & 0.28 & 0.26 & 0.26 & 2.71
& 0.27 & 0.32 & 0.31 & 0.28 & 0.28 & 0.32 & 4.81
& 0.34 & 0.36 & 0.36 & 0.33 & 0.33 & 0.37 & 6.98 \\
& \textbf{KVCMAS}
& 0.26 & 0.26 & 0.27 & 0.27 & 0.27 & 0.29 & 1.73
& 0.27 & 0.28 & 0.28 & 0.28 & 0.28 & 0.32 & 3.44
& 0.33 & 0.33 & 0.33 & 0.33 & 0.35 & 0.36 & 5.02 \\
\bottomrule
\end{tabular}
}

\vspace{5pt}

\textbf{(b) Shared generated context: 6k tokens}\\[2pt]
\resizebox{\textwidth}{!}{
\begin{tabular}{clrrrrrrrrrrrrrrrrrrrrr}
\toprule
& &
\multicolumn{7}{c}{$\rho=0.9$} &
\multicolumn{7}{c}{$\rho=0.8$} &
\multicolumn{7}{c}{$\rho=0.6$} \\
\cmidrule(lr){3-9}\cmidrule(lr){10-16}\cmidrule(lr){17-23}
\textbf{Pct.} & \textbf{Method}
& 0 & .25 & .5 & 1 & 2 & 4 & 8
& 0 & .25 & .5 & 1 & 2 & 4 & 8
& 0 & .25 & .5 & 1 & 2 & 4 & 8 \\
\midrule
p50 & NonShared
& 0.47 & 0.49 & 0.49 & 0.49 & 0.55 & 8.03 & 12.40
& 0.47 & 0.49 & 0.49 & 0.49 & 0.55 & 8.03 & 12.40
& 0.47 & 0.49 & 0.49 & 0.49 & 0.55 & 8.03 & 12.40 \\
& FullShared
& 0.19 & 0.19 & 0.19 & 0.19 & 0.19 & 0.18 & 2.92
& 0.19 & 0.19 & 0.19 & 0.19 & 0.19 & 0.18 & 2.92
& 0.19 & 0.19 & 0.19 & 0.19 & 0.19 & 0.18 & 2.92 \\
& DroidSpeak
& 0.37 & 0.38 & 0.38 & 0.36 & 0.41 & 5.73 & 9.25
& 0.37 & 0.38 & 0.36 & 0.36 & 0.41 & 5.67 & 9.32
& 0.37 & 0.38 & 0.38 & 0.36 & 0.40 & 5.46 & 9.21 \\
& CacheBlend
& 0.31 & 0.32 & 0.32 & 0.31 & 0.33 & 4.65 & 7.93
& 0.37 & 0.38 & 0.38 & 0.36 & 0.40 & 5.75 & 9.50
& 0.38 & 0.40 & 0.40 & 0.38 & 0.42 & 5.66 & 9.62 \\
& RelayCaching
& 0.32 & 0.34 & 0.33 & 0.32 & 0.34 & 4.54 & 7.96
& 0.37 & 0.39 & 0.38 & 0.36 & 0.41 & 5.83 & 9.55
& 0.38 & 0.40 & 0.40 & 0.38 & 0.43 & 5.74 & 9.66 \\
& GraphFlow
& 0.22 & 0.25 & 0.25 & 0.25 & 0.24 & 1.24 & 4.79
& 0.22 & 0.25 & 0.25 & 0.25 & 0.25 & 2.24 & 6.17
& 0.32 & 0.39 & 0.37 & 0.37 & 0.40 & 4.57 & 8.62 \\
& \textbf{KVCMAS}
& 0.19 & 0.19 & 0.20 & 0.21 & 0.22 & 0.29 & 4.04
& 0.22 & 0.24 & 0.25 & 0.25 & 0.25 & 1.21 & 5.42
& 0.32 & 0.33 & 0.33 & 0.35 & 0.39 & 3.99 & 8.12 \\
\midrule
p90 & NonShared
& 0.68 & 0.72 & 0.72 & 0.73 & 0.80 & 14.20 & 21.66
& 0.68 & 0.72 & 0.72 & 0.73 & 0.80 & 14.20 & 21.66
& 0.68 & 0.72 & 0.72 & 0.73 & 0.80 & 14.20 & 21.66 \\
& FullShared
& 0.23 & 0.25 & 0.25 & 0.25 & 0.25 & 0.25 & 4.46
& 0.23 & 0.25 & 0.25 & 0.25 & 0.25 & 0.25 & 4.46
& 0.23 & 0.25 & 0.25 & 0.25 & 0.25 & 0.25 & 4.46 \\
& DroidSpeak
& 0.57 & 0.62 & 0.61 & 0.62 & 0.62 & 9.51 & 16.51
& 0.57 & 0.63 & 0.61 & 0.62 & 0.62 & 9.50 & 16.64
& 0.57 & 0.62 & 0.61 & 0.61 & 0.62 & 9.20 & 16.43 \\
& CacheBlend
& 0.48 & 0.53 & 0.53 & 0.53 & 0.53 & 7.17 & 13.81
& 0.57 & 0.62 & 0.61 & 0.61 & 0.62 & 9.61 & 16.40
& 0.59 & 0.64 & 0.64 & 0.64 & 0.64 & 9.53 & 17.12 \\
& RelayCaching
& 0.49 & 0.53 & 0.53 & 0.53 & 0.53 & 7.26 & 14.02
& 0.57 & 0.64 & 0.62 & 0.62 & 0.62 & 9.61 & 16.49
& 0.59 & 0.66 & 0.64 & 0.64 & 0.64 & 9.65 & 17.20 \\
& GraphFlow
& 0.35 & 0.38 & 0.36 & 0.36 & 0.39 & 1.85 & 7.88
& 0.35 & 0.49 & 0.46 & 0.46 & 0.47 & 3.30 & 10.79
& 0.63 & 0.63 & 0.60 & 0.60 & 0.61 & 8.00 & 15.38 \\
& \textbf{KVCMAS}
& 0.35 & 0.37 & 0.37 & 0.37 & 0.39 & 0.66 & 6.42
& 0.35 & 0.39 & 0.40 & 0.41 & 0.47 & 1.93 & 9.10
& 0.61 & 0.61 & 0.61 & 0.62 & 0.63 & 6.80 & 14.22 \\
\bottomrule
\end{tabular}
}
\end{table*}

\newpage
\section{Ablation Studies}
\label{app:abl}

\subsection{Per-Request Efficiency}
\label{app:abl_static}

Section~\ref{sec:exp_eff} evaluates agent-request TTFT under concurrent serving.
Here, we measure memory and per-request efficiency under single-stream execution on an NVIDIA A6000 48\,GB GPU using Llama-3.1-8B, three agents, and 128 generated tokens per agent.
The workflow contains an initial dense-prefill request followed by two reuse requests.
TTFT is averaged over the reuse requests, while throughput divides the total trajectory tokens by E2E latency.

\paragraph{Memory Usage.}
Table~\ref{tab:static_mem_app} reports peak GPU memory across shared context lengths and reuse ratios.
KVComm exceeds the A6000 capacity because its full-dimensional anchor pool is stored alongside the model and active KV cache, so we estimate its requirement from the base caches and delta corrections.
At 2K shared tokens, KVComm runs out of memory due to memory fragmentation.
At 2K, 8K, and 32K, these estimates are 2.5$\times$, 6.2$\times$, and 14.9$\times$ those of KVCMAS, respectively.
KVCMAS maintains peak memory comparable to the selective recomputation methods and nearly unchanged across reuse ratios  because it maintains a fixed number of low-rank anchors rather than full-dimensional recomputation states.

\begin{table*}[h]
\centering
\caption{
Peak GPU memory in GB under single-stream execution across shared context lengths and reuse ratios.
KVComm$^\dagger$ exceeds the A6000 capacity, so its requirement is estimated from its full-dimensional anchor states.
}
\label{tab:static_mem_app}
\setlength{\tabcolsep}{6pt}
\begin{tabular}{clrrrrr}
\toprule
\textbf{Shared context}
& \textbf{Method}
& \multicolumn{5}{c}{\textbf{Reuse ratio $\rho$}} \\
\cmidrule(lr){3-7}
& & \textbf{0.95} & \textbf{0.90} & \textbf{0.80}
& \textbf{0.70} & \textbf{0.60} \\
\midrule
\multirow{8}{*}{2K}
& NonShared       & 16.9 & 16.9 & 16.9 & 16.9 & 16.9 \\
& FullShared      & 16.4 & 16.4 & 16.4 & 16.4 & 16.4 \\
& DroidSpeak      & 17.0 & 17.0 & 17.0 & 17.1 & 17.2 \\
& CacheBlend      & 17.2 & 17.2 & 17.2 & 17.3 & 17.4 \\
& RelayCaching    & 17.4 & 17.4 & 17.4 & 17.5 & 17.5 \\
& KVComm$^\dagger$ & 42.3 & 42.3 & 42.3 & 42.3 & 42.3 \\
& GraphFlow       & 16.8 & 16.8 & 16.8 & 16.8 & 16.8 \\
& \textbf{KVCMAS} & \textbf{17.1} & \textbf{17.1} & \textbf{17.1} & \textbf{17.1} & \textbf{17.1} \\
\midrule
\multirow{8}{*}{8K}
& NonShared       & 19.8 & 19.8 & 19.8 & 19.8 & 19.8 \\
& FullShared      & 17.9 & 17.9 & 17.9 & 17.9 & 17.9 \\
& DroidSpeak      & 19.3 & 19.3 & 19.5 & 19.8 & 20.0 \\
& CacheBlend      & 19.4 & 19.5 & 19.7 & 19.9 & 20.1 \\
& RelayCaching    & 19.6 & 19.7 & 19.8 & 19.9 & 20.0 \\
& KVComm$^\dagger$ & 120.9 & 120.9 & 120.9 & 120.9 & 120.9 \\
& GraphFlow       & 19.1 & 19.1 & 19.1 & 19.1 & 19.1 \\
& \textbf{KVCMAS} & \textbf{19.5} & \textbf{19.5} & \textbf{19.5} & \textbf{19.5} & \textbf{19.5} \\
\midrule
\multirow{8}{*}{32K}
& NonShared       & 31.5 & 31.5 & 31.5 & 31.5 & 31.5 \\
& FullShared      & 24.0 & 24.0 & 24.0 & 24.0 & 24.0 \\
& DroidSpeak      & 28.6 & 28.7 & 29.5 & 30.6 & 31.4 \\
& CacheBlend      & 28.5 & 28.9 & 29.7 & 30.5 & 31.3 \\
& RelayCaching    & 28.5 & 28.7 & 29.1 & 29.5 & 29.9 \\
& KVComm$^\dagger$ & 435.6 & 435.6 & 435.6 & 435.6 & 435.6 \\
& GraphFlow       & 28.9 & 28.9 & 28.9 & 28.9 & 28.9 \\
& \textbf{KVCMAS} & \textbf{29.2} & \textbf{29.2} & \textbf{29.2} & \textbf{29.2} & \textbf{29.2} \\
\bottomrule
\end{tabular}
\end{table*}

\begin{figure*}[h]
    \centering
    \vspace{-20pt}
    \includegraphics[width=\textwidth]{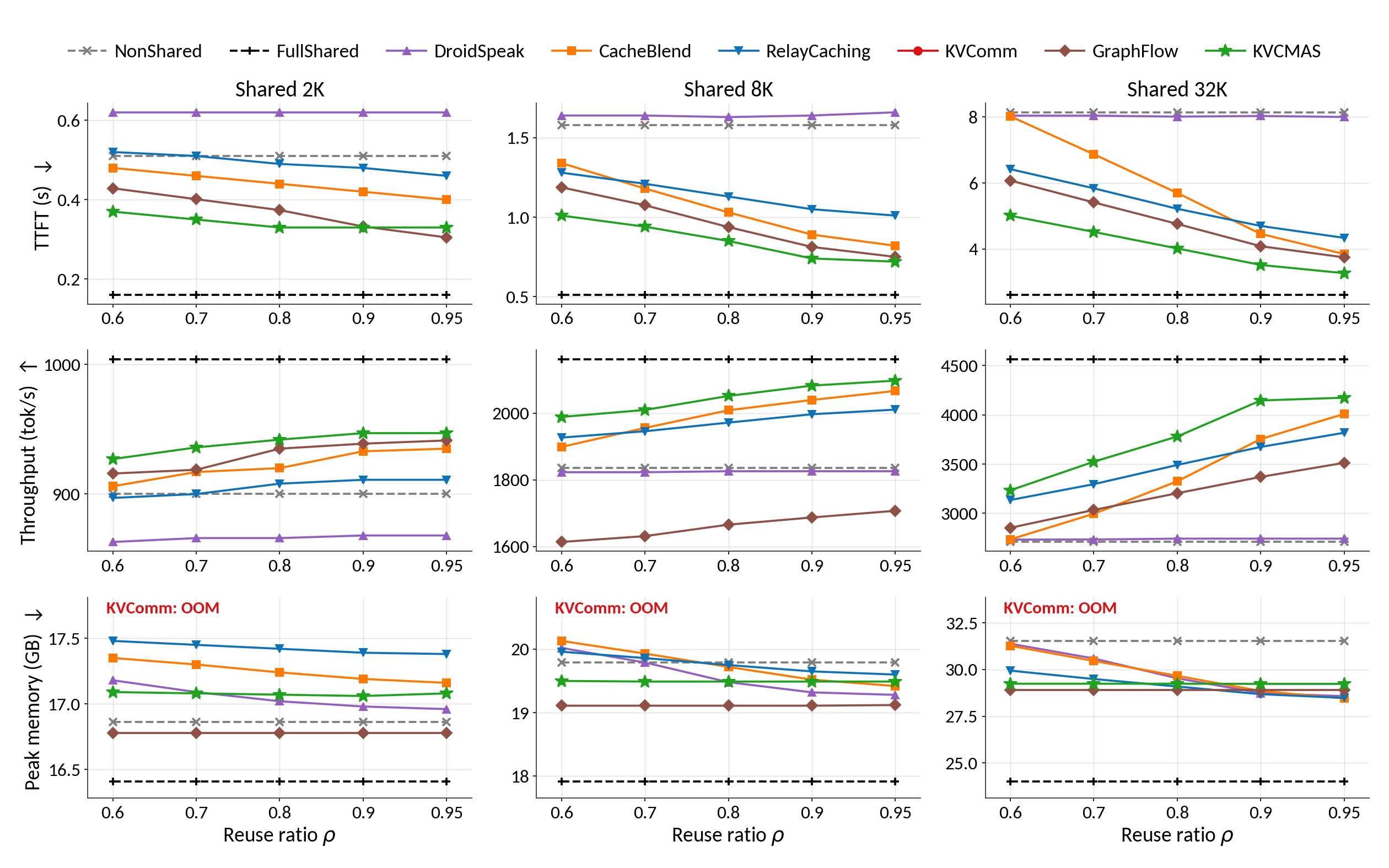}
    \vspace{-20pt}
    \caption{
    Single-stream TTFT, E2E throughput, and peak GPU memory across shared context lengths at matched reuse ratios.
    Lower TTFT and memory and higher throughput indicate better efficiency.
    KVComm exceeds the NVIDIA A6000 48\,GB memory capacity.
    }
    \label{fig:static_eff_app}
\end{figure*}

\paragraph{TTFT and Throughput.}
Table~\ref{tab:static_eff_app} reports the numerical results, while Figure~\ref{fig:static_eff_app} compares the methods at matched reuse ratios.
At 8K and 32K, KVCMAS achieves the lowest TTFT and highest throughput among the KV cache sharing works, with larger gains as the shared context grows.
At 2K, where prefill contributes less to execution, it remains comparable to the fastest work.
These results demonstrate its efficiency in single-batch serving, including settings relevant to edge deployments, particularly for context-heavy workloads.

\newpage

\renewcommand{\arraystretch}{0.98}
\begin{table*}[h]
\centering
\caption{
Average reuse-request TTFT and E2E trajectory throughput under single-stream execution.
TTFT is reported in seconds, and throughput in tokens/s.
}
\label{tab:static_eff_app}
\setlength{\tabcolsep}{3.2pt}
\resizebox{\textwidth}{!}{
\begin{tabular}{clrrrrrrrrrr}
\toprule
&
&
\multicolumn{5}{c}{\textbf{TTFT (s)}}
&
\multicolumn{5}{c}{\textbf{Throughput (tokens/s)}} \\
\cmidrule(lr){3-7}\cmidrule(lr){8-12}
\textbf{Shared context}
& \textbf{Method}
& \multicolumn{5}{c}{$\rho$}
& \multicolumn{5}{c}{$\rho$} \\
\cmidrule(lr){3-7}\cmidrule(lr){8-12}
& & \textbf{0.95} & \textbf{0.90} & \textbf{0.80} & \textbf{0.70} & \textbf{0.60}
& \textbf{0.95} & \textbf{0.90} & \textbf{0.80} & \textbf{0.70} & \textbf{0.60} \\
\midrule
\multirow{7}{*}{2K}
& NonShared
& 0.51 & 0.51 & 0.51 & 0.51 & 0.51
& 900 & 900 & 900 & 900 & 900 \\
& FullShared
& 0.16 & 0.16 & 0.16 & 0.16 & 0.16
& 1,004 & 1,004 & 1,004 & 1,004 & 1,004 \\
& DroidSpeak
& 0.62 & 0.62 & 0.62 & 0.62 & 0.62
& 868 & 868 & 866 & 866 & 863 \\
& CacheBlend
& 0.40 & 0.42 & 0.44 & 0.46 & 0.48
& 935 & 933 & 920 & 917 & 906 \\
& RelayCaching
& 0.46 & 0.48 & 0.49 & 0.51 & 0.52
& 911 & 911 & 908 & 900 & 897 \\
& GraphFlow
& 0.31 & 0.33 & 0.37 & 0.40 & 0.43
& 941 & 939 & 935 & 919 & 916 \\
& \textbf{KVCMAS}
& \textbf{0.33} & \textbf{0.33} & \textbf{0.33} & \textbf{0.35} & \textbf{0.37}
& \textbf{947} & \textbf{947} & \textbf{942} & \textbf{936} & \textbf{927} \\
\midrule
\multirow{7}{*}{8K}
& NonShared
& 1.58 & 1.58 & 1.58 & 1.58 & 1.58
& 1,836 & 1,836 & 1,836 & 1,836 & 1,836 \\
& FullShared
& 0.51 & 0.51 & 0.51 & 0.51 & 0.51
& 2,162 & 2,162 & 2,162 & 2,162 & 2,162 \\
& DroidSpeak
& 1.66 & 1.64 & 1.63 & 1.64 & 1.64
& 1,826 & 1,826 & 1,826 & 1,823 & 1,823 \\
& CacheBlend
& 0.82 & 0.89 & 1.03 & 1.18 & 1.34
& 2,067 & 2,040 & 2,009 & 1,956 & 1,899 \\
& RelayCaching
& 1.01 & 1.05 & 1.13 & 1.21 & 1.28
& 2,011 & 1,997 & 1,972 & 1,946 & 1,927 \\
& GraphFlow
& 0.75 & 0.81 & 0.94 & 1.08 & 1.19
& 1,707 & 1,687 & 1,666 & 1,631 & 1,614 \\
& \textbf{KVCMAS}
& \textbf{0.72} & \textbf{0.74} & \textbf{0.85} & \textbf{0.94} & \textbf{1.01}
& \textbf{2,098} & \textbf{2,083} & \textbf{2,052} & \textbf{2,010} & \textbf{1,989} \\
\midrule
\multirow{7}{*}{32K}
& NonShared
& 8.14 & 8.14 & 8.14 & 8.14 & 8.14
& 2,709 & 2,709 & 2,709 & 2,709 & 2,709 \\
& FullShared
& 2.61 & 2.61 & 2.61 & 2.61 & 2.61
& 4,564 & 4,564 & 4,564 & 4,564 & 4,564 \\
& DroidSpeak
& 8.00 & 8.03 & 8.01 & 8.04 & 8.04
& 2,743 & 2,743 & 2,743 & 2,734 & 2,733 \\
& CacheBlend
& 3.85 & 4.46 & 5.70 & 6.87 & 8.02
& 4,009 & 3,754 & 3,325 & 2,995 & 2,735 \\
& RelayCaching
& 4.34 & 4.70 & 5.22 & 5.84 & 6.42
& 3,818 & 3,674 & 3,491 & 3,296 & 3,135 \\
& GraphFlow
& 3.75 & 4.09 & 4.76 & 5.41 & 6.08
& 3,513 & 3,370 & 3,204 & 3,034 & 2,853 \\
& \textbf{KVCMAS}
& \textbf{3.27} & \textbf{3.52} & \textbf{4.02} & \textbf{4.52} & \textbf{5.01}
& \textbf{4,174} & \textbf{4,146} & \textbf{3,780} & \textbf{3,524} & \textbf{3,234} \\
\bottomrule
\end{tabular}
}
\end{table*}
\renewcommand{\arraystretch}{1.0}

\newpage
\subsection{Ranks of Base Cache and Delta Correction}
\label{app:abl_rank}

\begin{figure*}[h]
    \centering
    \includegraphics[width=\textwidth]{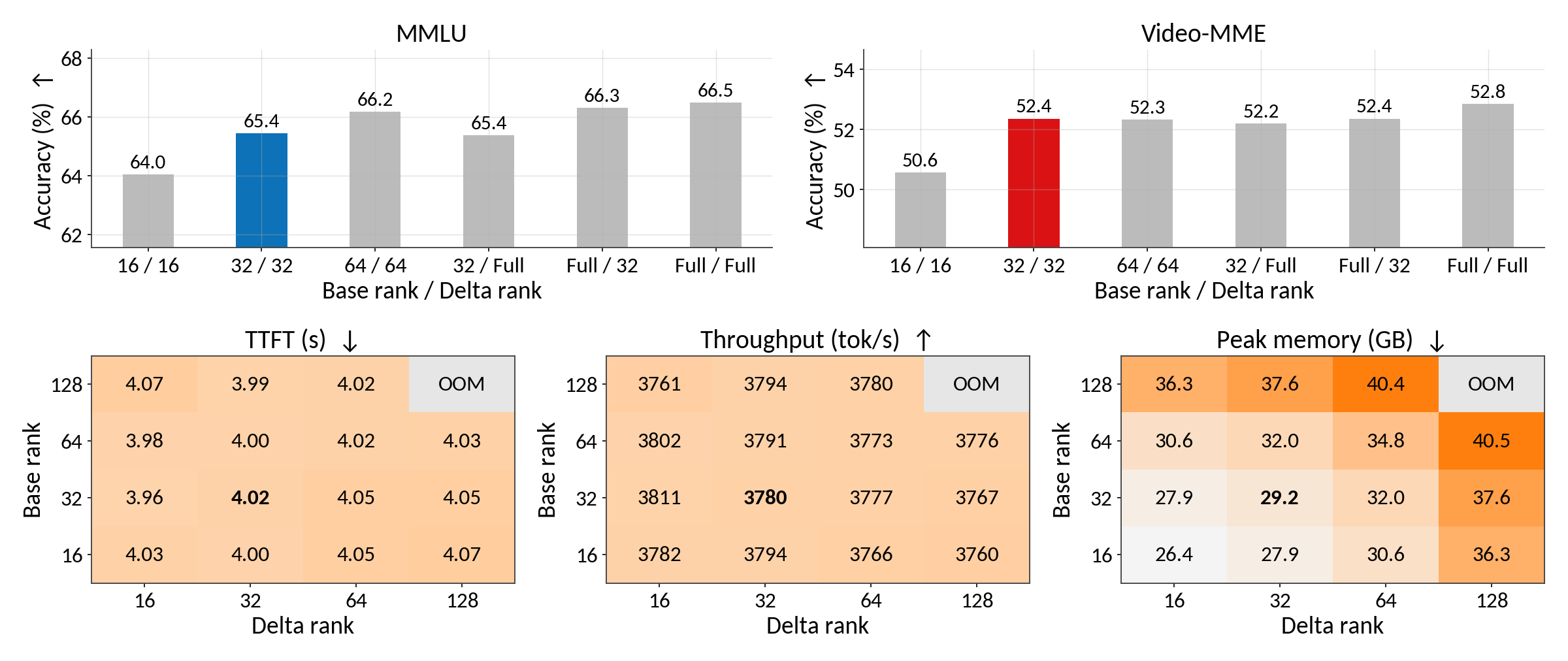}
    \vspace{-20pt}
\caption{
Effect of the base cache and delta correction ranks on accuracy and single-stream efficiency.
The upper panels report accuracy on MMLU and Video-MME, while the lower panels show TTFT, throughput, and peak GPU memory across rank combinations.
``Full'' denotes the corresponding uncompressed KV feature dimension, and the default $(r_C,r_{\Delta})=(32,32)$ is highlighted.
}
    \label{fig:abl_rank}
\end{figure*}

Section~\ref{sec:method_lr} represents the source cache and delta correction using low-rank factors with ranks $r_C$ and $r_{\Delta}$, respectively.
We vary them independently and evaluate accuracy on MMLU and Video-MME using their benchmark-specific reliability thresholds, adjusting $\tau$ for each rank configuration to maintain the target reuse regime of around $\rho=0.8$ and $0.6$ for the LLM and VLM tasks, respectively.
In addition, we measure TTFT, throughput, and peak GPU memory with Llama-3.1-8B under the 32K single-stream setting with three agents and $\rho=0.8$.
Figure~\ref{fig:abl_rank} reports the accuracy and efficiency results, where ``Full'' denotes the corresponding uncompressed KV feature dimension.

Increasing both ranks from 16 to 32 improves accuracy on MMLU and Video-MME.
Beyond rank 32, accuracy changes marginally while memory usage continues to increase, and the $(128,128)$ configuration exceeds GPU capacity.
TTFT and throughput remain nearly unchanged among configurations that fit in memory.
We therefore use $r_C=r_{\Delta}=32$, which captures most of the full-dimensional accuracy while maintaining a compact anchor pool.

\newpage
\subsection{Anchor Pool Size}
\label{app:abl_anchor}

KVCMAS maintains a process-wide anchor pool for each shared placeholder slot.
Each anchor stores a low-rank base cache and consumer-specific corrections, and its capacity $V$ is applied independently to each slot.
While Section~\ref{sec:method_kvcmas} describes one source--target transition for clarity, the implementation shares these slot-specific pools across requests.
We vary $V$ under the single-stream setting with 32K shared tokens, three agents, rank $r=32$, and $\rho=0.8$.
KVComm uses $V=20$ as its original balance between accuracy and efficiency, while KVCMAS uses $V=10$ by default. 
For accuracy, we adjust $\tau$ for each pool capacity to maintain approximately $\rho=0.8$ on MMLU and $\rho=0.6$ on Video-MME.
Because entropy-based reliability requires at least two compatible anchors, the $V=1$ configuration bypasses the reliability gate and directly uses the sole anchor.

\begin{table}[h]
\centering
\caption{
Effect of anchor pool capacity $V$ on single-stream efficiency and accuracy.
TTFT is reported in seconds, throughput in tokens/s, peak GPU memory in GB, and accuracy in percent.
}
\label{tab:abl_anchor}
\begin{tabular}{crrrrr}
\toprule
\textbf{$V$}
& \textbf{TTFT (s)}
& \textbf{Throughput}
& \textbf{Memory}
& \textbf{MMLU}
& \textbf{Video-MME} \\
\midrule
1  & 2.69 & 4,050 & 26.21 & 62.51 & 49.40 \\
5  & 3.93 & 3,821 & 26.68 & 64.42 & 51.89 \\
10 & 4.02 & 3,780 & 29.24 & 65.45 & 52.36 \\
20 & 4.05 & 3,769 & 31.94 & 65.80 & 53.15 \\
\bottomrule
\end{tabular}
\end{table}

A larger pool provides more representative anchors and improves correction accuracy, with $V=10$ capturing most of the gain.
Increasing $V$ from 10 to 20 provides only marginal additional accuracy with higher memory usage, while TTFT and throughput change little beyond $V=5$.
Even at the matched capacity of $V=20$, KVCMAS uses 31.94\,GB, compared with the estimated 435.6\,GB required by KVComm under the same 32K configuration, isolating the low-rank memory reduction from the difference in their default pool capacities.
We therefore use $V=10$ as a balanced default.

\newpage
\subsection{Scaling with the Number of Agents}
\label{app:abl_agentn}

Using Llama-3.1-8B on an NVIDIA A6000 48\,GB GPU, we vary the number of agents $N\in\{2,4,8\}$ with 32K shared tokens, one interaction round, and 128 generated tokens per agent under single-stream execution.
TTFT is averaged across all $N$ agent requests, while throughput covers the full trajectory.
Selective recomputation uses $\rho=0.9$, and delta correction uses $\rho=0.8$.
Because the fixed 32K context dominates the trajectory length, increasing $N$ primarily creates more opportunities to reuse the same context.

\begin{table*}[h]
\centering
\caption{
Single-stream efficiency as the number of agents increases.
Selective recomputation uses $\rho=0.9$, and delta correction uses $\rho=0.8$.
KVComm is omitted because its anchor pool exceeds GPU capacity at 32K shared tokens.
OOM denotes out of memory.
}
\label{tab:abl_agentn}
\resizebox{\textwidth}{!}{
\begin{tabular}{lrrrrrrrrr}
\toprule
& \multicolumn{3}{c}{\textbf{TTFT (s)}}
& \multicolumn{3}{c}{\textbf{Throughput (token/s)}}
& \multicolumn{3}{c}{\textbf{Memory (GB)}} \\
\cmidrule(lr){2-4}\cmidrule(lr){5-7}\cmidrule(lr){8-10}
\textbf{Method}
& \textbf{$N=2$} & \textbf{$N=4$} & \textbf{$N=8$}
& \textbf{$N=2$} & \textbf{$N=4$} & \textbf{$N=8$}
& \textbf{$N=2$} & \textbf{$N=4$} & \textbf{$N=8$} \\
\midrule
NonShared
& 8.25 & 8.37 & \textsc{oom}
& 2,729 & 2,760 & \textsc{oom}
& 27.25 & 35.63 & \textsc{oom} \\
FullShared
& 3.87 & 2.01 & 1.08
& 3,830 & 4,922 & 5,904
& 23.06 & 23.06 & 23.06 \\
DroidSpeak
& 6.82 & 6.19 & 5.91
& 2,657 & 2,852 & 3,021
& 27.12 & 27.93 & 29.64 \\
CacheBlend
& 5.63 & 4.49 & 3.90
& \textbf{3,627} & 3,900 & 4,153
& 27.03 & 27.79 & 29.54 \\
RelayCaching
& 5.53 & 4.28 & 3.65
& 3,480 & 3,883 & 4,093
& 26.61 & 27.22 & 28.44 \\
GraphFlow
& 5.71 & 3.97 & 3.10
& 3,075 & 3,860 & 4,244
& 28.80 & 29.82 & 31.86 \\
\textbf{KVCMAS}
& \textbf{5.11} & \textbf{3.50} & \textbf{2.70}
& 3,546 & \textbf{4,327} & \textbf{4,776}
& 29.07 & 29.39 & 30.03 \\
\bottomrule
\end{tabular}
}
\end{table*}

KVCMAS achieves the lowest TTFT among the KV cache sharing works across all agent counts, with a larger advantage as more agents reuse the shared context.
It also provides the highest throughput at $N=4$ and $N=8$ and remains close to the highest at $N=2$.
More agents amortize the initial dense prefill across additional reuse requests, reducing average TTFT.
Peak memory increases only marginally because corrections for additional agent transitions are stored as compact low-rank anchor states.

\newpage
\subsection{Scaling with the Number of Rounds}
\label{app:abl_round}

Using Llama-3.1-8B on an NVIDIA A6000 48\,GB GPU, we vary the number of interaction rounds $R\in\{2,4,8\}$ with three agents, 32K initial shared tokens, and 128 generated tokens per agent under single-stream execution.
TTFT is averaged across all agent requests, while throughput covers the full trajectory.
The fixed 32K context remains dominant as additional rounds create more opportunities to reuse the accumulated context.
Selective recomputation uses $\rho=0.9$, and delta correction uses $\rho=0.8$.

\begin{table*}[h]
\centering
\caption{
Single-stream efficiency as the number of interaction rounds increases.
Selective recomputation uses $\rho=0.9$, and delta correction uses $\rho=0.8$.
KVComm is omitted because its anchor pool exceeds GPU capacity at 32K shared tokens.
}
\label{tab:abl_round}
\resizebox{\textwidth}{!}{
\begin{tabular}{lrrrrrrrrr}
\toprule
& \multicolumn{3}{c}{\textbf{TTFT (s)}}
& \multicolumn{3}{c}{\textbf{Throughput (token/s)}}
& \multicolumn{3}{c}{\textbf{Memory (GB)}} \\
\cmidrule(lr){2-4}\cmidrule(lr){5-7}\cmidrule(lr){8-10}
\textbf{Method}
& \textbf{$R=2$} & \textbf{$R=4$} & \textbf{$R=8$}
& \textbf{$R=2$} & \textbf{$R=4$} & \textbf{$R=8$}
& \textbf{$R=2$} & \textbf{$R=4$} & \textbf{$R=8$} \\
\midrule
NonShared
& 8.22 & 8.33 & 8.19
& 2,728 & 2,747 & 2,769
& 31.44 & 31.44 & 31.44 \\
FullShared
& 1.39 & 0.74 & 0.43
& 5,421 & 6,445 & 6,948
& 23.06 & 23.06 & 23.06 \\
DroidSpeak
& 5.87 & 5.61 & 5.54
& 2,756 & 2,777 & 2,805
& 27.66 & 27.75 & 28.03 \\
CacheBlend
& 4.03 & 3.57 & 3.36
& 3,927 & 4,110 & 4,278
& 27.50 & 27.65 & 27.90 \\
RelayCaching
& 3.75 & 3.37 & 3.08
& 3,824 & 3,972 & 4,138
& 27.00 & 27.16 & 27.38 \\
GraphFlow
& 3.04 & 2.48 & 2.21
& 3,926 & 4,248 & 4,582
& 29.36 & 29.52 & 29.70 \\
\textbf{KVCMAS}
& \textbf{2.87} & \textbf{2.30} & \textbf{2.02}
& \textbf{4,198} & \textbf{4,485} & \textbf{4,802}
& 29.33 & 29.49 & 29.77 \\
\bottomrule
\end{tabular}
}
\end{table*}

KVCMAS achieves the lowest TTFT and highest throughput among the KV cache sharing works across all evaluated round counts.
Additional rounds amortize the initial dense prefill across more reuse requests, reducing average TTFT and increasing throughput.
Peak GPU memory increases slightly because the growing trajectory adds compact low-rank anchor states for newly accumulated agent outputs.

\end{document}

%% file: math_commands.tex
\usepackage{amsmath,amsfonts,bm}

\def\eqref#1{equation~\ref{#1}}

\def\1{\bm{1}}

\DeclareMathAlphabet{\mathsfit}{\encodingdefault}{\sfdefault}{m}{sl}
\SetMathAlphabet{\mathsfit}{bold}{\encodingdefault}{\sfdefault}{bx}{n}

